\PassOptionsToPackage{table}{xcolor}

\documentclass[manuscript, screen]{acmart}
\AtBeginDocument{%
  }

\setcopyright{none}
\renewcommand\footnotetextcopyrightpermission[1]{}

\usepackage{multirow}
\usepackage{makecell}

\usepackage{tcolorbox}
\usepackage{enumitem}
\usepackage{wrapfig}
\usepackage{ulem}
\usepackage{placeins}

\microtypesetup{expansion=false}

\definecolor{myblue}{HTML}{D2E6F1}
\definecolor{mypink}{HTML}{FDCCCC}
\definecolor{thered}{RGB}{247,105,108}   
\definecolor{thegreen}{RGB}{99,190,123}  
\definecolor{myred}{HTML}{D3393C}
\definecolor{mygreen}{HTML}{3A9D5D}
\definecolor{revblue}{HTML}{0066CC}  

\newcommand{\hl}[1]{#1} 

\tcbset{mybox/.style={
  colframe=black,
  colback=white,
  boxrule=0.5mm,
  left=2mm,
  right=2mm,
  top=2mm,
  bottom=2mm,
  width=\textwidth,
  enlarge left by=-2mm,
  enlarge right by=-2mm
}}

\newcommand{\colorcell}[1]{%
  \pgfmathparse{#1}%
  \ifdim\pgfmathresult pt<0pt
    \pgfmathsetmacro{\val}{min(100, abs(#1)/30 * 100)} 
    \edef\colorval{\noexpand\cellcolor{thered!\val!white}}%
  \else
    \pgfmathsetmacro{\val}{min(100, #1/631.4 * 100)} 
    \edef\colorval{\noexpand\cellcolor{thegreen!\val!white}}%
  \fi
  \colorval
}

\newcommand{\calimp}[1]{%
    \ifdim#1 pt > 0pt %
        \textcolor{mygreen}{\tiny ↑\the\numexpr #1\relax\%}%
    \else
        \textcolor{myred}{\tiny ↓\the\numexpr -#1\relax\%}%
    \fi
}

\newcommand{\valuewithoutimp}[2]{%
    \textnormal{#1}\textsubscript{\tiny$\pm$#2}
}

\begin{document}

\title{Large Language Models as Virtual Survey Respondents: Evaluating Sociodemographic Response Generation}

\author{Jianpeng Zhao}
\affiliation{%
  \institution{University of Macau}
  \city{Macau}
  \country{China}}
\email{zhaojianpengcs@gmail.com}

\author{Chenyu Yuan}
\affiliation{%
  \institution{University of Macau}
  \city{Macau}
  \country{China}}

\author{Weiming Luo}
\affiliation{%
  \institution{University of Macau}
  \city{Macau}
  \country{China}
}

\author{Haoling Xie}
\affiliation{%
 \institution{The Chinese University of Hong Kong}
 \city{Hong Kong}
 \country{China}}

\author{Guangwei Zhang}
\affiliation{%
  \institution{City University of Hong Kong}
  \city{Hong Kong}
  \country{China}}

\author{Steven Jige Quan}
\affiliation{%
  \institution{Seoul National University}
  \city{Seoul}
  \country{South Korea}
}

\author{Zixuan Yuan}
\affiliation{%
  \institution{The Hong Kong University of Science and Technology (Guangzhou)}
  \city{Guangzhou}
  \state{Guangdong}
  \country{China}}

\author{Pengyang Wang}
\affiliation{%
  \institution{University of Macau}
  \city{Macau}
  \country{China}}

\author{Denghui Zhang}
\affiliation{%
  \institution{Stevens Institute of Technology}
  \city{Hoboken}
  \state{New Jersey}
  \country{USA}
}

\renewcommand{\shortauthors}{Jianpeng Zhao et al.}

\begin{abstract}
Questionnaire-based surveys are foundational to social science research and public policymaking, yet traditional survey methods remain costly, time-consuming, and often limited in scale.
\hl{Although prior work has explored large language models (LLMs) as virtual survey respondents, existing studies often address narrow task settings, focus on single sociological domains, or lack a unified evaluation framework that enables systematic comparison across diverse datasets and models.}
\hl{To address these gaps, we introduce two complementary task abstractions: \textbf{Partial Attribute Simulation (PAS)}, where LLMs predict missing attributes from incomplete respondent profiles, and \textbf{Full Attribute Simulation (FAS)}, where LLMs generate complete synthetic datasets under zero-context and context-enhanced conditions. Both are framed as diagnostic and exploratory tools rather than replacements for human data collection.}
\hl{We curate \textbf{LLM-S\textsuperscript{3}} (\underline{\textbf{L}}arge \underline{\textbf{L}}anguage \underline{\textbf{M}}odel-based \underline{\textbf{S}}ociodemographic \underline{\textbf{S}}urvey \underline{\textbf{S}}imulation), a benchmark spanning 11 real-world public datasets across four sociological domains, and evaluate GPT-3.5/4 Turbo and LLaMA 3.0/3.1-8B under zero-shot and few-shot settings.}
Our evaluation reveals consistent performance trends across model families, highlights failure modes in structured output generation, and demonstrates how context and prompt design affect simulation fidelity.
Our code and dataset are available at:
\url{https://github.com/dart-lab-research/LLM-S-Cube-Benchmark}.

\end{abstract}



\maketitle

\section{Introduction}
\label{sec:introduction}

Questionnaire-based surveys serve as critical instruments for understanding public sentiment, tracking social dynamics, and guiding evidence-based policymaking.
However, the traditional survey pipeline faces mounting challenges: high operational costs, lengthy data collection timelines, and declining response rates~\cite{survey_from_Individual_to_society}.
These limitations hinder the timely delivery of insights essential for responsive policy formulation.
\hl{To address these issues, we investigate a structured framework for leveraging Large Language Models (LLMs) in survey response generation.}

\begin{figure}[h!]
\centering
\includegraphics[width=1\textwidth]{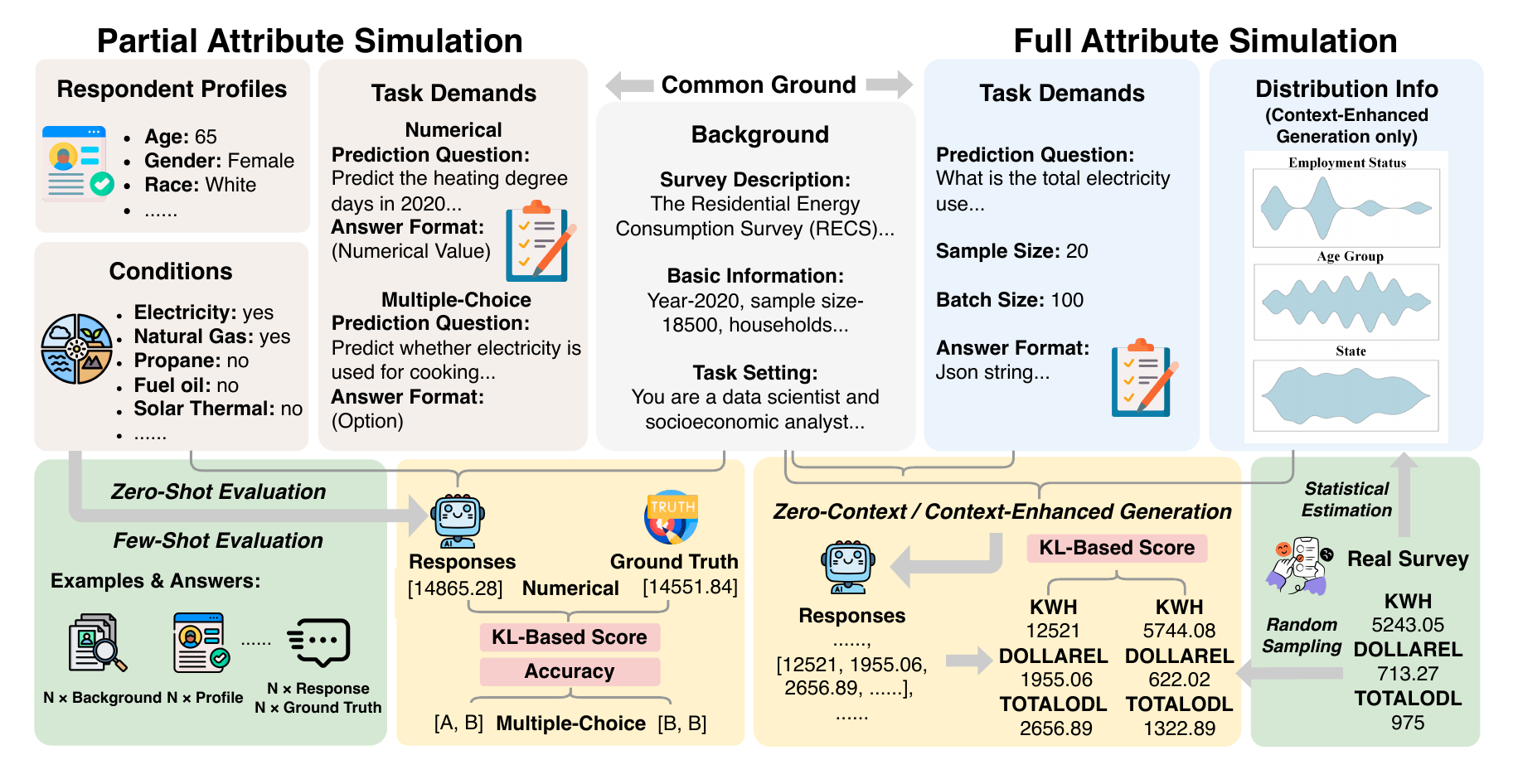}
\caption{An Overview of Partial Attribute Simulation (left) and Full Attribute Simulation (right).}
\label{fig:framework}
\end{figure}

Recent advances in LLMs have demonstrated remarkable generalization capabilities in tasks ranging from question answering to synthetic text generation~\cite{gpt3,llama3,tornberg2023chatgpt}.
\hl{While there is growing interest in LLM-driven simulations~\cite{dillion2023can,kim2023ai,jansen2023employing} and prior work has explored LLMs as survey respondents~\cite{argyle2023out,park2024generative,bisbee2024synthetic}, existing research often suffers from three limitations that our work addresses:} (1) narrow task coverage, where prior studies frequently focus on a limited number of survey items or demographic traits, restricting their generalizability across diverse tasks~\cite{llm_predictors,ChatGPT_vs_Social_Surveys, Random_Silicon_Sampling,argyle2023out};
(2) domain-specific constraints, as most existing works are tailored to particular sociological themes (e.g., politics or consumer behavior), which impedes comprehensive cross-domain evaluation~\cite{Large_language_models_as_instruments_of_power,trump,argyle2023out,tornberg2023simulating,cao-etal-2025-specializing}; and
(3) inadequate benchmarking, primarily due to reliance on synthetic or overly simplified datasets that fail to reflect the complexity or scale of real-world survey distributions~\cite{argyle2023out,ziems2024can,ren2024valuebench,cao-etal-2025-specializing}.
Our research bridges these gaps by introducing a systematic and extensible framework for evaluating LLMs in survey response generation.

To this end, we introduce two distinct and complementary task settings: \textbf{Partial Attribute Simulation (PAS)}, where LLMs predict missing demographic or behavioral attributes from incomplete respondent profiles, and
\textbf{Full Attribute Simulation (FAS)}, where LLMs generate synthetic datasets under varying degrees of contextual constraint.
These paradigms are designed to reflect practical scenarios in applied research.
\hl{For instance, PAS addresses settings where a government agency possesses detailed demographic records ({\it e.g.}, age, income, occupation) for its residents but lacks information on specific opinions or preferences, such as attitudes toward public transportation or renewable energy. By conditioning on the known attributes, PAS can generate plausible predictions for the missing survey responses, enabling population-level trend analysis without conducting a full-scale survey.
In contrast, FAS targets scenarios where an organization, such as a market research firm, has access to aggregate statistics about a target population ({\it e.g.}, age distributions and income brackets) but cannot obtain individual-level profiles. FAS can then synthesize a statistically representative synthetic population, supporting exploratory analyses such as consumer preference estimation or policy impact assessment.}
\hl{We clarify that PAS and FAS are best understood as diagnostic, methodological, or exploratory tools rather than a viable replacement for human survey data collection.}

To support reproducibility and broad applicability, we curate a cross-domain benchmark suite,
\textbf{LLM-S\textsuperscript{3}}
(\underline{\textbf{L}}arge \underline{\textbf{L}}anguage \underline{\textbf{M}}odel-based \underline{\textbf{S}}ociodemographic \underline{\textbf{S}}urvey \underline{\textbf{S}}imulation),
composed of eleven public datasets across four sociological areas: Social and Public Affairs, Work and Income, Household and Behavioral Patterns, and Health and Lifestyle.
\hl{We note that the datasets used in this study are predominantly drawn from North American and European contexts, reflecting the current availability of large-scale public survey data. While this scope is a practical constraint of the present evaluation rather than a fundamental limitation of the PAS/FAS framework itself, the benchmark is designed to be extensible: as cross-cultural and multilingual survey datasets become available, the same evaluation protocol can be applied to assess LLM performance across diverse sociological contexts.}
We evaluate four representative LLMs (GPT-3.5 Turbo, GPT-4 Turbo, LLaMA 3.0-8B, LLaMA 3.1-8B) on both numerical and categorical prediction tasks under zero-shot and few-shot settings.
Our extensive experiments yield several key findings: (i) LLMs exhibit reliable and consistent performance trends across various model families and task types;
(ii) prompt design and contextual augmentation significantly impact simulation fidelity, highlighting their crucial role; and
(iii) failure cases in structured output generation remain a major bottleneck, particularly evident in the Full Attribute Simulation (FAS) scenario.
\hl{We explicitly acknowledge that our evaluation measures statistical and distributional similarity, not behavioral fidelity, construct validity, or the nuanced reasoning of real human respondents.}
These insights not only advance our fundamental understanding of LLMs in survey applications but also provide practical design principles for future research and deployment in this emerging field.


\section{Related Work}
\label{sec:related_work}

\subsection{\hl{From Role-Playing to Structured Survey Simulation}}

Recent advancements in LLMs have enabled increasingly realistic agent behaviors through persona-driven role-playing, where maintaining coherent persona traits and consistent decision-making is essential \cite{peng2024quantifying, wang2024incharacter, xu2024character}. User-specific information such as traits and profiles has been incorporated into dialogue agents to enhance personalization \cite{zhang-etal-2018-personalizing, lamp}, and evaluation frameworks have been developed to measure how well LLMs replicate human-like personalized responses \cite{llm_echo_us, PersonalityChat}. These efforts demonstrate that LLMs can effectively simulate human behavior when grounded in appropriate contextual and demographic information.

\hl{The natural extension of this capability is to apply LLM role-playing to sociological research, using models as virtual survey respondents. However, a critical mismatch emerges between the two domains. Role-playing and persona-faithfulness studies~\cite{peng2024quantifying, wang2024incharacter, xu2024character} are designed for open-domain dialogue or narrative settings, where the goal is to maintain character consistency over extended interactions. Survey simulation, by contrast, requires structured, attribute-conditioned prediction: the model must infer a specific Likert-scale response given a partial demographic profile, without the benefit of conversational context or extended character development. The metrics that work for dialogue consistency (persona faithfulness, interview fidelity) do not transfer to the task of predicting survey responses from structured demographic inputs. This fundamental mismatch motivates the need for a distinct framework that formalizes survey simulation as a rigorous, reproducible task with its own evaluation criteria.}

\subsection{\hl{Survey Simulation: From Feasibility to Methodological Rigor}}

\hl{Research on LLM-based survey simulation has progressed through three distinct phases, each revealing both capabilities and critical limitations that shape our understanding of what is required for reliable simulation.}

\hl{\textbf{Phase 1: Establishing Feasibility.} Early work demonstrated that LLMs can be prompted to reproduce broad attitudinal patterns across demographic subgroups. Argyle et al.~\cite{argyle2023out} showed that models encode sufficient sociological knowledge to simulate survey responses, establishing the ``silicon sampling'' paradigm. While this established basic feasibility, it also raised immediate questions: What are the conditions under which LLM responses reliably track human distributions? What factors limit generalizability?}

\hl{\textbf{Phase 2: Pushing Toward Behavioral Realism.} Subsequent work addressed these questions by grounding simulations in richer individual-level data. Park et al.~\cite{park2024generative} constructed detailed agent memories from interview data, enabling virtual respondents to exhibit behavioral consistency over time. This approach demonstrated that individual-level grounding improves simulation quality, but the requirement of extensive per-individual data collection limits scalability and practical applicability.}

\hl{\textbf{Phase 3: Evaluating Validity.} The most recent work has introduced methodological rigor by explicitly testing whether LLM-generated data can substitute for human responses in actual research workflows. Bisbee et al.~\cite{bisbee2024synthetic} found that even when distributions appear statistically similar, they can lead to substantively different inferential conclusions. This finding is pivotal: it establishes that distributional similarity is necessary but not sufficient for valid survey simulation, and that the gap between statistical fidelity and behavioral fidelity is an open methodological challenge requiring systematic investigation.}

\hl{\textbf{Summary: Persistent Structural Gaps.} While these three phases have advanced our understanding of LLM capabilities in survey simulation, the field has not converged on a shared conceptual framework or evaluation standard. This fragmentation leads to three specific structural gaps: (1) \textbf{lack of conceptual clarity} about what tasks survey simulation should perform (predicting missing attributes from partial profiles versus generating full synthetic populations from aggregate statistics); (2) \textbf{no multi-domain benchmark} enabling consistent evaluation across diverse sociological settings; and (3) \textbf{no standardized evaluation protocols}, hindering fair comparison across studies.}

\hl{Our work addresses these gaps directly. The PAS/FAS abstraction brings conceptual clarity to two distinct simulation regimes: partial versus full attribute simulation. The LLM-S\textsuperscript{3} benchmark provides a unified, multi-domain evaluation suite spanning eleven real-world datasets across four sociological domains, enabling systematic cross-domain comparison that prior work has not supported.}

\section{Partial Attribute Simulation (PAS)}
\label{sec:section3}
\hl{In numerous real-world survey scenarios, an organization already holds structured demographic records for a population of interest but lacks the specific attitudinal or behavioral measurements it needs. For example, a government labor agency might maintain administrative records for thousands of employees, including age, gender, job level, education, and employment type, yet lack survey data on job satisfaction, stress, or work-life balance. Conducting a full survey is costly and time-consuming; however, the agency still needs distributional estimates of these outcomes to inform workforce policy. PAS is designed for precisely this setting: the available demographic attributes serve as the input profile, while the unmeasured attitudinal traits act as the prediction targets. Conditioned on each individual's known profile, the LLM is prompted to predict the missing responses, enabling population-level trend analysis without a full-scale survey. Crucially, PAS does not require any individual-level response history. It relies solely on the demographic data that the organization already possesses, which distinguishes it from collaborative filtering or imputation methods that require at least partial response data for the target attributes.}
As illustrated in Figure~\ref{fig:framework}, PAS operates by providing LLMs with only fundamental demographic data and selected conditional attributes, without access to an individual's complete historical response profile.
The LLMs are then tasked with generating plausible responses, effectively simulating how individuals with specific partial characteristics might answer survey questions.
This paradigm offers valuable insights into the influence of partial information on survey responses, enabling the simulation of realistic survey scenarios across diverse sociological contexts.

\subsection{Formulation}
Within the Partial Attribute Simulation (PAS) framework, survey respondents are conceptualized as virtual agents, each characterized by a predefined set of attributes. 
Let $\mathcal{D} = \{\mathbf{x}_1, \mathbf{x}_{2},..., \mathbf{x}_N\}$ denote a survey dataset of $N$ samples, where each $\mathbf{x}_i \in \mathbb{R}^M$ consists of $M$ attributes from the attribute set $\mathcal{A}$. 
The set $\mathcal{A}$ is partitioned into two distinct subsets: the input attribute set, $\mathcal{A}_{\text{prior}}$, which are provided to the Large Language Model (LLM) via prompts, and the target attribute set, $\mathcal{A}_{\text{target}}$, that the LLM is required to infer. Formally, we define:

\begin{equation}
      \mathcal{A}_{\text{prior}} \subseteq \mathcal{A}, 
\mathcal{A}_{\text{target}} = \mathcal{A} \setminus \mathcal{A}_{\text{prior}}.
\end{equation}

Each virtual agent is then tasked with predicting the target attributes $\mathbf{A}_{\text{target}} \in \mathcal{A}_{\text{target}}$ conditioned on the given input attributes $\mathbf{A}_{\text{prior}} \in \mathcal{A}_{\text{prior}}$. 
This is formalized as modeling the conditional probability distribution parameterized by $\theta$:
\begin{equation}
    \mathbf{A}_{\text{target}} \sim P_{\theta}(\mathbf{A}_{\text{target}}|\mathbf{A}_{\text{prior}}). 
\end{equation}
 
In the PAS scenario, each respondent is simulated as an independent virtual agent using an LLM, which is required to generate a distribution over the possible values of $\mathbf{A}_{\text{target}}$ conditioned on $\mathbf{A}_{\text{prior}}$.

\textbf{Numerical Prediction Tasks:} When predicting numerical attributes, the primary objective is to estimate the population-level distribution of the target attribute set $\mathcal{A}_{\text{target}}$, rather than predicting specific values for individual respondents. 
Performance is assessed using a KL-based score (see Appendix~\ref{appendix:numerical_pred_task}), a monotonic decreasing transformation of the Kullback–Leibler (KL) divergence, where higher values indicate better alignment between predicted and ground-truth distributions. 

\textbf{Multiple-Choice Prediction Tasks:} For categorical attributes, the focus shifts from modeling distributions to determining the correct category for each individual respondent. 
In this context, the LLM's performance is evaluated based on the accuracy of its predictions for each respondent's target attribute, with the average accuracy across all respondents being reported. 
This metric directly assesses the model's classification capability. 

\subsection{Evaluation Strategies in PAS}
In the PAS scenario, two primary evaluation strategies are employed to assess LLM performance: \textbf{Zero-Shot Evaluation} and \textbf{Few-Shot Evaluation}.

In the \textbf{Zero-Shot Evaluation}, the model generates predictions solely based on its pre-trained parametric knowledge, without any prior examples. The prompt, detailed in Figure~\ref{fig:prompt_design_pas} in Appendix~\ref{sec:appendix_A}, is structured around five key components:

\begin{itemize}
    \item \textbf{Background:} Consists of a survey description (purpose and structure of the questionnaire), basic information (outlining context such as year, region, domain, etc.) and a task setting description (instructions guiding the LLM to simulate a specific respondent).
    \item \textbf{Respondent Profiles:} Includes demographic information ({\it e.g.}, age, gender) to tailor predictions to individual characteristics. 
    \item \textbf{Conditions:} Provides conditional attributes ({\it e.g.}, electricity usage, health factors) to guide predictions based on specific circumstances. 
    \item \textbf{Task Demands:} Specifies the prediction task type ({\it e.g.}, numerical or multiple-choice).
    \item \textbf{Answer Format:} Defines the output format ({\it e.g.}, numerical/option lists, or JSON strings).
\end{itemize}

In the \textbf{Few-Shot Evaluation}, the model refines its predictions of $\mathcal{A}_{target}$ by learning from a small set of in-context examples~\cite{ICL}. These exemplars act as demonstrations, helping the model recognize the expected response format and improve predictive accuracy without explicit task-specific training. 
In this study, the LLM receives five illustrative examples, providing structural guidance and reinforcing expected patterns. 
See Figure~\ref{fig:prompt_design_pas} in Appendix~\ref{sec:appendix_A} for a prompt example. 

\textbf{Experimental Setup:} Based on these two evaluation strategies, we evaluated public datasets from four domains (see Table~\ref{tab:dataset-desciption}) with key attributes selected for prediction (see Table~\ref{tab:pas_attri}). 
Numerical attributes were evaluated using a KL-based performance score (refer to Appendix~\ref{appendix:numerical_pred_task} for calculation details), while multiple-choice tasks were assessed by accuracy. 
Performance across both tasks, evaluated using OpenAI's GPT-3.5 Turbo, GPT-4 Turbo~\cite{gpt3}, and Meta's LLaMA 3.0 and 3.1 (8B) models~\cite{llama3}, is presented in Table~\ref{tab:pas_perf_numer} and~\ref{tab:pas_perf_mc}. 
GPT models were accessed via their official API\footnote{\url{https://openai.com/api/}}, and LLaMA models were deployed locally using the Ollama platform\footnote{\url{https://ollama.ai/}} on an NVIDIA A6000 GPU, with default hyperparameters applied for unbiased comparisons.

\begin{table}[htbp]
  \centering
  \small
  \setlength{\tabcolsep}{2.25pt}
  \renewcommand{\arraystretch}{1.25}
  \caption{LLM performance on numerical prediction in PAS. The KL-based scores (↑) are reported as {\it avg.~$\pm$~std.}: bold for the best and underlined for the second-best.
  }
  \label{tab:pas_perf_numer}
  \resizebox{\textwidth}{!}{
  \begin{tabular}{l | cccc | cccc}
    \Xhline{1.2pt}
    \multirow{2}{*}{\textbf{Dataset}} &
    \multicolumn{4}{c|}{\textbf{Zero-Shot Evaluation}} &
    \multicolumn{4}{c}{\textbf{Few-Shot Evaluation}} \\
    \cline{2-5} \cline{6-9}
    & GPT-3.5 Turbo & GPT-4 Turbo & LLaMA 3.0 & LLaMA 3.1 & GPT-3.5 Turbo & GPT-4 Turbo & LLaMA 3.0 & LLaMA 3.1 \\
    \hline
    ANES & \valuewithoutimp{\textbf{0.5735}}{0.0529} & \valuewithoutimp{\uline{0.5576}}{0.0293} & \valuewithoutimp{0.4875}{0.0533} & \valuewithoutimp{0.4768}{0.0755}
         & \valuewithoutimp{\textbf{0.5568}}{0.0072}
         & \valuewithoutimp{\uline{0.5460}}{0.0494}
         & \valuewithoutimp{0.4646}{0.0575}
         & \valuewithoutimp{0.4683}{0.0334} \\
    GSS & \valuewithoutimp{0.6983}{0.0120} & \valuewithoutimp{\textbf{0.7082}}{0.0129} & \valuewithoutimp{\uline{0.6990}}{0.0198} & \valuewithoutimp{0.6818}{0.0062}
        & \valuewithoutimp{0.6766}{0.0086}
        & \valuewithoutimp{\textbf{0.7054}}{0.0164}
        & \valuewithoutimp{\uline{0.6879}}{0.0142}
        & \valuewithoutimp{0.6800}{0.0308} \\
    BIS & \valuewithoutimp{\textbf{0.6204}}{0.0462} & \valuewithoutimp{0.5119}{0.0129} & \valuewithoutimp{\uline{0.5777}}{0.0458} & \valuewithoutimp{0.5730}{0.0378}
        & \valuewithoutimp{\textbf{0.6207}}{0.0417}
        & \valuewithoutimp{0.5299}{0.0337}
        & \valuewithoutimp{\uline{0.5640}}{0.0486}
        & \valuewithoutimp{0.5622}{0.0476} \\
    RECS & \valuewithoutimp{\textbf{0.6554}}{0.0162} & \valuewithoutimp{\uline{0.5648}}{0.0568} & \valuewithoutimp{0.3114}{0.0801} & \valuewithoutimp{0.2669}{0.0197}
         & \valuewithoutimp{\uline{0.5293}}{0.1865}
         & \valuewithoutimp{\textbf{0.5489}}{0.0920}
         & \valuewithoutimp{0.3088}{0.0874}
         & \valuewithoutimp{0.2365}{0.0226} \\
    ACS & \valuewithoutimp{0.5159}{0.1283} & \valuewithoutimp{0.4722}{0.1859} & \valuewithoutimp{\textbf{0.6056}}{0.0983} & \valuewithoutimp{\uline{0.5766}}{0.1239}
        & \valuewithoutimp{0.4791}{0.1106}
        & \valuewithoutimp{0.3636}{0.0410}
        & \valuewithoutimp{\textbf{0.6054}}{0.1047}
        & \valuewithoutimp{\uline{0.5969}}{0.1112} \\
    Trell SMU & \valuewithoutimp{0.3523}{0.0340} & \valuewithoutimp{0.4301}{0.0033} & \valuewithoutimp{\textbf{0.5612}}{0.0003} & \valuewithoutimp{\uline{0.5361}}{0.0064}
              & \valuewithoutimp{0.4274}{0.0499}
              & \valuewithoutimp{\uline{0.5197}}{0.0670}
              & \valuewithoutimp{\textbf{0.5462}}{0.0189}
              & \valuewithoutimp{0.4685}{0.0147} \\
    NHTS & \valuewithoutimp{0.4010}{0.0066} & \valuewithoutimp{0.3125}{0.0842} & \valuewithoutimp{\uline{0.4927}}{0.1769} & \valuewithoutimp{\textbf{0.6079}}{0.1036}
         & \valuewithoutimp{0.3568}{0.0997}
         & \valuewithoutimp{0.3102}{0.1499}
         & \valuewithoutimp{\uline{0.4908}}{0.1058}
         & \valuewithoutimp{\textbf{0.5799}}{0.0760} \\
    \Xhline{1.2pt}
  \end{tabular}
  }
\end{table}

\begin{table}[htbp]
  \centering
  \small
  \setlength{\tabcolsep}{2.25pt}
  \renewcommand{\arraystretch}{1.25}
  \caption{LLM performance on multiple-choice prediction in PAS. 
  Accuracy (↑) is reported as avg.~$\pm$~std.: bold for the best and underlined for the second-best. 
  Cell color reflects relative performance change compared to random-choice accuracy: greener for greater improvement and redder for greater degradation.}

  \label{tab:pas_perf_mc}
  \resizebox{\textwidth}{!}{
  \begin{tabular}{l | cccc | cccc}
    \Xhline{1.2pt}
    \multirow{2}{*}{\textbf{Dataset}} &
    \multicolumn{4}{c|}{\textbf{Zero-Shot Evaluation}} &  
    \multicolumn{4}{c}{\textbf{Few-Shot Evaluation}} \\
    \cline{2-5} \cline{6-9}
    & GPT-3.5 Turbo & GPT-4 Turbo & LLaMA 3.0 & LLaMA 3.1 & 
    GPT-3.5 Turbo & GPT-4 Turbo & LLaMA 3.0 & LLaMA 3.1 \\
    \hline
    ANES
      & \colorcell{57.08}\valuewithoutimp{\uline{0.7854}}{0.0598}
      & \colorcell{67.16}\valuewithoutimp{\textbf{0.8358}}{0.0324}
      & \colorcell{56.54}\valuewithoutimp{0.7827}{0.0487}
      & \colorcell{50.50}\valuewithoutimp{0.7525}{0.0372}
      & \colorcell{55.92}\valuewithoutimp{\uline{0.7796}}{0.0447}
      & \colorcell{64.38}\valuewithoutimp{\textbf{0.8219}}{0.0395}
      & \colorcell{53.72}\valuewithoutimp{0.7686}{0.0382}
      & \colorcell{47.10}\valuewithoutimp{0.7355}{0.0491} \\
    GSS
      & \colorcell{179.40}\valuewithoutimp{\uline{0.5588}}{0.1359}
      & \colorcell{215.65}\valuewithoutimp{\textbf{0.6313}}{0.0340}
      & \colorcell{164.45}\valuewithoutimp{0.5289}{0.0583}
      & \colorcell{148.45}\valuewithoutimp{0.4969}{0.0326}
      & \colorcell{147.40}\valuewithoutimp{\uline{0.4948}}{0.0430}
      & \colorcell{211.15}\valuewithoutimp{\textbf{0.6223}}{0.0349}
      & \colorcell{146.65}\valuewithoutimp{0.4933}{0.0895}
      & \colorcell{106.45}\valuewithoutimp{0.4129}{0.0778} \\
    BIS
      & \colorcell{-6.75}\valuewithoutimp{0.1865}{0.0032}
      & \colorcell{8.25}\valuewithoutimp{0.2165}{0.0022}
      & \colorcell{31.45}\valuewithoutimp{\uline{0.2629}}{0.0600}
      & \colorcell{42.40}\valuewithoutimp{\textbf{0.2848}}{0.0217}
      & \colorcell{-26.75}\valuewithoutimp{0.1465}{0.0164}
      & \colorcell{8.45} \valuewithoutimp{\uline{0.2169}}{0.0045}
      & \colorcell{13.05}\valuewithoutimp{\textbf{0.2261}}{0.0233}
      & \colorcell{3.85}\valuewithoutimp{0.2077}{0.0044} \\
    RECS
      & \colorcell{24.16}\valuewithoutimp{0.6208}{0.1116}
      & \colorcell{71.96}\valuewithoutimp{\textbf{0.8598}}{0.0793}
      & \colorcell{45.14}\valuewithoutimp{\uline{0.7257}}{0.0144}
      & \colorcell{30.80}\valuewithoutimp{0.6540}{0.0730}
      & \colorcell{27.72}\valuewithoutimp{0.6386}{0.0257}
      & \colorcell{82.26}\valuewithoutimp{\textbf{0.9113}}{0.0165}
      & \colorcell{49.66}\valuewithoutimp{\uline{0.7483}}{0.2132}
      & \colorcell{34.84}\valuewithoutimp{0.6742}{0.0623} \\
    EmpS
      & \colorcell{9.55}\valuewithoutimp{\uline{0.2191}}{0.0293}
      & \colorcell{8.55}\valuewithoutimp{0.2171}{0.0017}
      & \colorcell{15.10}\valuewithoutimp{\textbf{0.2302}}{0.0259}
      & \colorcell{9.40}\valuewithoutimp{0.2188}{0.0183}
      & \colorcell{13.35}\valuewithoutimp{\textbf{0.2267}}{0.0092}
      & \colorcell{5.80}\valuewithoutimp{0.2116}{0.0105}
      & \colorcell{7.35}\valuewithoutimp{0.2147}{0.0230}
      & \colorcell{7.15}\valuewithoutimp{0.2143}{0.0045} \\
    NHTS
      & \colorcell{631.40}\valuewithoutimp{\textbf{0.7314}}{0.0077}
      & \colorcell{623.10}\valuewithoutimp{\uline{0.7231}}{0.0003}
      & \colorcell{473.20}\valuewithoutimp{0.5732}{0.0126}
      & \colorcell{550.70}\valuewithoutimp{0.6507}{0.0453}
      & \colorcell{624.60}\valuewithoutimp{\uline{0.7246}}{0.0322}
      & \colorcell{630.00}\valuewithoutimp{\textbf{0.7300}}{0.0057}
      & \colorcell{540.90}\valuewithoutimp{0.6409}{0.0797}
      & \colorcell{498.70}\valuewithoutimp{0.5987}{0.0270} \\
    Trell SMU
      & \colorcell{284.90}\valuewithoutimp{\uline{0.6415}}{0.0105}
      & \colorcell{291.50}\valuewithoutimp{\textbf{0.6525}}{0.0050}
      & \colorcell{131.90}\valuewithoutimp{0.3865}{0.0138}
      & \colorcell{201.68}\valuewithoutimp{0.5028}{0.0134}
      & \colorcell{333.68}\valuewithoutimp{\uline{0.7228}}{0.0357}
      & \colorcell{296.06}\valuewithoutimp{0.6601}{0.0057}
      & \colorcell{325.94}\valuewithoutimp{0.7099}{0.0051}
      & \colorcell{338.54}\valuewithoutimp{\textbf{0.7309}}{0.0176} \\
    MxMH
      & \colorcell{41.13}\valuewithoutimp{\textbf{0.1283}}{0.0380}
      & \colorcell{12.31}\valuewithoutimp{\uline{0.1021}}{0.0018}
      & \colorcell{-0.89}\valuewithoutimp{0.0901}{0.0129}
      & \colorcell{2.63}\valuewithoutimp{0.0933}{0.0151}
      & \colorcell{69.62}\valuewithoutimp{\textbf{0.1542}}{0.0517}
      & \colorcell{7.47}\valuewithoutimp{\uline{0.0977}}{0.0093}
      & \colorcell{1.20}\valuewithoutimp{0.0920}{0.0103}
      & \colorcell{-5.40}\valuewithoutimp{0.0860}{0.0120} \\
    YPS
      & \colorcell{36.45}\valuewithoutimp{0.2729}{0.0074}
      & \colorcell{27.95}\valuewithoutimp{0.2559}{0.0038}
      & \colorcell{54.15}\valuewithoutimp{\textbf{0.3083}}{0.0479}
      & \colorcell{39.35}\valuewithoutimp{\uline{0.2787}}{0.0426}
      & \colorcell{45.80}\valuewithoutimp{\uline{0.2916}}{0.0283}
      & \colorcell{48.40}\valuewithoutimp{\textbf{0.2968}}{0.0471}
      & \colorcell{38.40}\valuewithoutimp{0.2768}{0.0175}
      & \colorcell{-0.50}\valuewithoutimp{0.1990}{0.0418} \\
    MHD
      & \colorcell{1.31}\valuewithoutimp{0.3377}{0.0058}
      & \colorcell{1.25}\valuewithoutimp{0.3375}{0.0045}
      & \colorcell{1.70}\valuewithoutimp{\uline{0.3390}}{0.0068}
      & \colorcell{3.02}\valuewithoutimp{\textbf{0.3434}}{0.0070}
      & \colorcell{-0.58}\valuewithoutimp{0.3314}{0.0198}
      & \colorcell{0.17}\valuewithoutimp{0.3339}{0.0079}
      & \colorcell{2.09}\valuewithoutimp{\uline{0.3403}}{0.0094}
      & \colorcell{3.11}\valuewithoutimp{\textbf{0.3437}}{0.0063} \\
    \Xhline{1.2pt}
  \end{tabular}
  }
\end{table}

\subsection{PAS Performance Analysis}
\begin{figure}[htbp]
\centering
\begin{minipage}{0.7\textwidth}
    \centering
    \includegraphics[width=\linewidth]{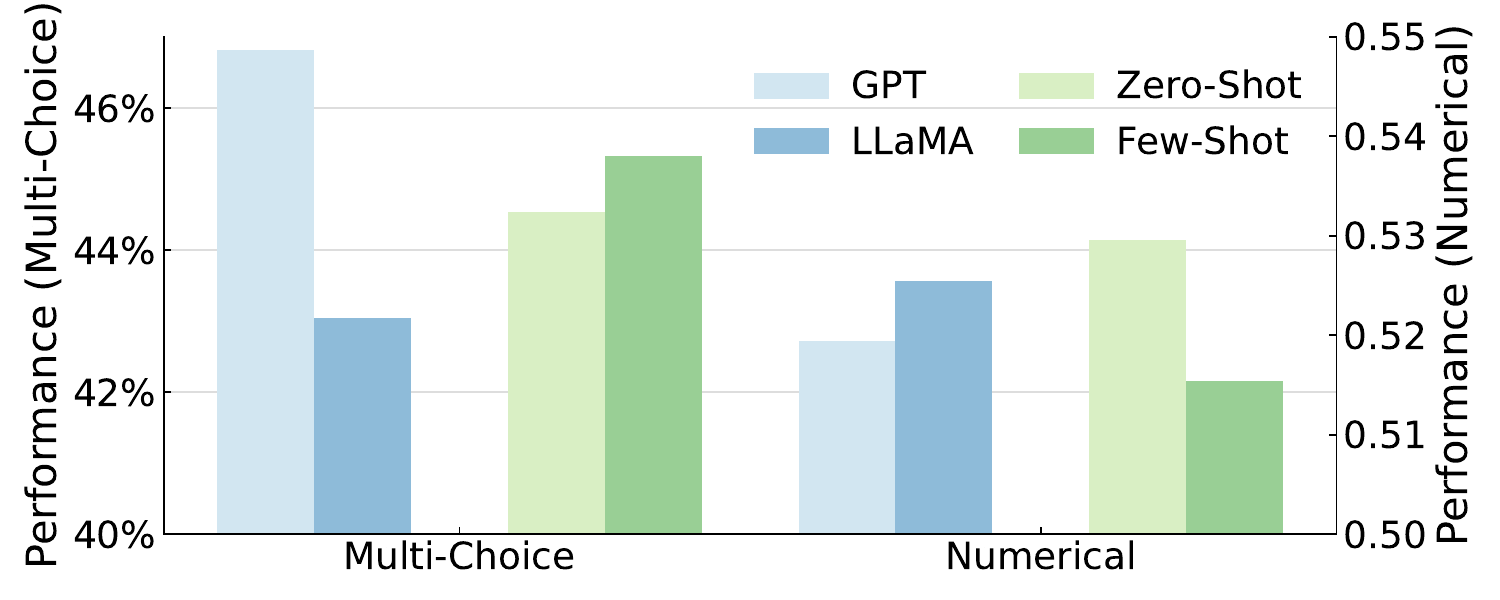}
\end{minipage}
\caption{Average PAS performance (↑) across two tasks, grouped by model series and evaluation strategy.}
\label{fig:pas_bar}
\end{figure}

To better understand how model series and evaluation strategies affect performance across different task types, we examine aggregate trends in the PAS scenario. 
As shown in Figure~\ref{fig:pas_bar}, the GPT series generally outperforms LLaMA on multiple-choice tasks (46.82\% {\it vs.} 43.04\%), while LLaMA achieves slightly better results on numerical prediction (0.5255 {\it vs.} 0.5194). 
Few-shot evaluation tends to improve performance on multiple-choice tasks (45.32\% {\it vs.} 44.54\%), whereas zero-shot proves more effective for numerical tasks (0.5296 {\it vs.} 0.5154). 
These divergent patterns highlight the need for a more granular, task-specific analysis. 

\textbf{Numerical Prediction Tasks:} Table~\ref{tab:pas_perf_numer} reveals that GPT models achieve the best performance on four datasets, while LLaMA leads on three. 
Although model series exhibit consistent performance trends, the effect of evaluation strategy is more variable. 
Few-shot settings improve performance in some cases ({\it e.g.}, GPT-4 Turbo on Trell SMU), but degrade it in others ({\it e.g.}, on ACS), underscoring its sensitivity to dataset properties and prompt design (as further discussed in Section~\ref{sec:few_shot_bias}). 

\textbf{Multiple-Choice Prediction Tasks:}
As presented in Table~\ref{tab:pas_perf_mc}, the majority of models outperform the random-choice accuracy, with particularly pronounced improvements observed on the NHTS and Trell SMU datasets.
GPT models, particularly GPT-4 Turbo, generally attain the highest accuracy across most datasets, with LLaMA outperforming them on a few datasets.
Consistent with numerical tasks, our few-shot evaluation exhibits variable improvements in this prediction task as well.

\subsection{\hl{Attribute Predictability and Demographic Profile Alignment in PAS}}
\hl{A central question for PAS is not merely whether an LLM can predict survey responses, but rather \textit{which attributes are predictable from a given demographic profile}. Aggregate accuracy alone cannot answer this question, as it conflates attributes that are strongly determined by observable demographics with those governed by latent psychological or contextual factors. To surface this distinction, we conduct a subgroup-level analysis on the EmpS dataset ($N{=}3{,}025$) using LLaMA~3.1-8B in the zero-shot setting. We select EmpS because its five target attributes (WLB, WorkEnv, Workload, Stress, JobSatisfaction) all measure related workplace attitude constructs on a uniform 5-level ordinal scale, and its input profile includes rich, directly relevant demographic covariates (age, gender, job level, department, employment type, education), enabling controlled cross-attribute comparison where differences in predictability can be attributed to the target construct itself rather than to variation in response scales, domains, or input feature relevance. We use LLaMA~3.1-8B because, as a fully open-weight model, it permits local deployment, deterministic decoding, and reproducible repeated runs without dependence on proprietary API availability or versioning changes. We note that this analysis is a focused diagnostic case study rather than a comprehensive cross-model evaluation. Prediction accuracy is disaggregated by age group, gender, and job level across three independent runs (reported as mean$\pm$std). All five target attributes are 5-choice questions, yielding a random baseline accuracy of 20\%.}

\begin{figure}[htbp]
\centering
\includegraphics[width=\textwidth]{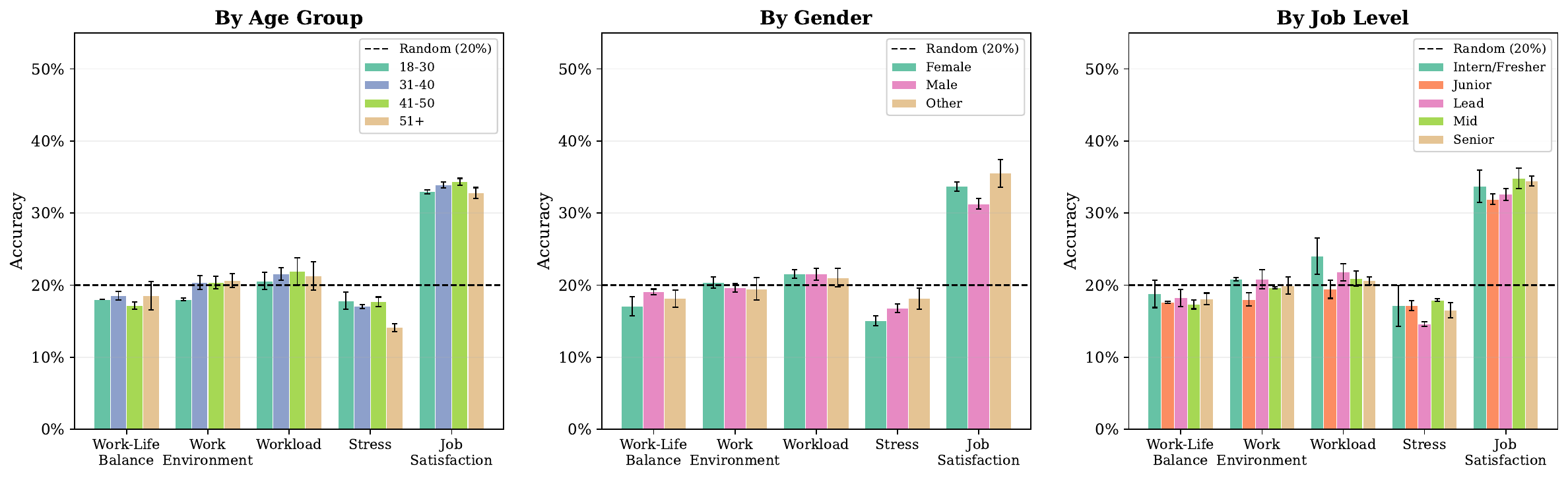}
\caption{\hl{PAS subgroup-level accuracy on the EmpS dataset (LLaMA 3.1-8B, zero-shot, $N{=}3{,}025$, averaged over 3 independent runs). Each bar represents mean prediction accuracy for a subgroup on a target attribute; error bars denote standard deviation across runs. The dashed line marks the random baseline (20\%). JobSatisfaction is the only attribute where all subgroups consistently exceed the baseline, while Stress falls below it across all demographic dimensions.}}
\label{fig:subgroup_bars}
\end{figure}

\hl{Figure~\ref{fig:subgroup_bars} presents the per-subgroup prediction accuracy. As reported in Table~\ref{tab:pas_perf_mc}, the overall mean accuracy of LLaMA~3.1 on EmpS under zero-shot evaluation is $21.9\%$, only marginally above the random baseline of $20\%$. This result is itself informative: it reveals that the EmpS demographic profile (age, gender, job level, education, employment type) provides limited predictive signal for most workplace attitude attributes, suggesting that these outcomes are shaped primarily by individual experience and organizational context rather than by sociodemographic characteristics alone. Critically, however, the picture is not uniform across attributes. JobSatisfaction ($33.5\%$) consistently exceeds the baseline across all subgroups and demographic dimensions, indicating that this attribute maintains a detectable statistical association with the provided profile features. Stress ($16.7\%$), by contrast, falls below the baseline, implying that stress levels are largely orthogonal to the demographic covariates available in EmpS. This attribute-level divergence constitutes a key insight enabled by the PAS benchmark: it allows practitioners to identify, prior to deployment, which survey constructs are amenable to demographic-profile-based simulation and which are not.}

\hl{The subgroup breakdown further refines this picture. For Stress, the 51+ age group achieves only $14.1{\pm}0.6\%$ accuracy, a $3.7$ percentage-point gap below the 18--30 cohort ($17.8{\pm}1.2\%$), and Lead-level employees are the hardest to predict ($14.6{\pm}0.3\%$), falling $3.3$ percentage points below the Mid-level group ($17.9{\pm}0.2\%$). These patterns suggest that stress responses among senior-age and leadership-role employees are particularly idiosyncratic, likely reflecting heterogeneous organizational pressures that demographic proxies cannot capture. For JobSatisfaction, the gender dimension reveals a $4.2$ percentage-point gap between the Other category ($35.5{\pm}1.9\%$) and Male respondents ($31.3{\pm}0.8\%$), consistent with the hypothesis that satisfaction correlates more strongly with demographic identity for non-binary respondents in this dataset. Across all subgroups, the low variance across runs (std ${\leq}2.8\%$) confirms that these patterns are stable and not artifacts of sampling noise.}

\begin{figure}[htbp]
\centering
\includegraphics[width=\textwidth]{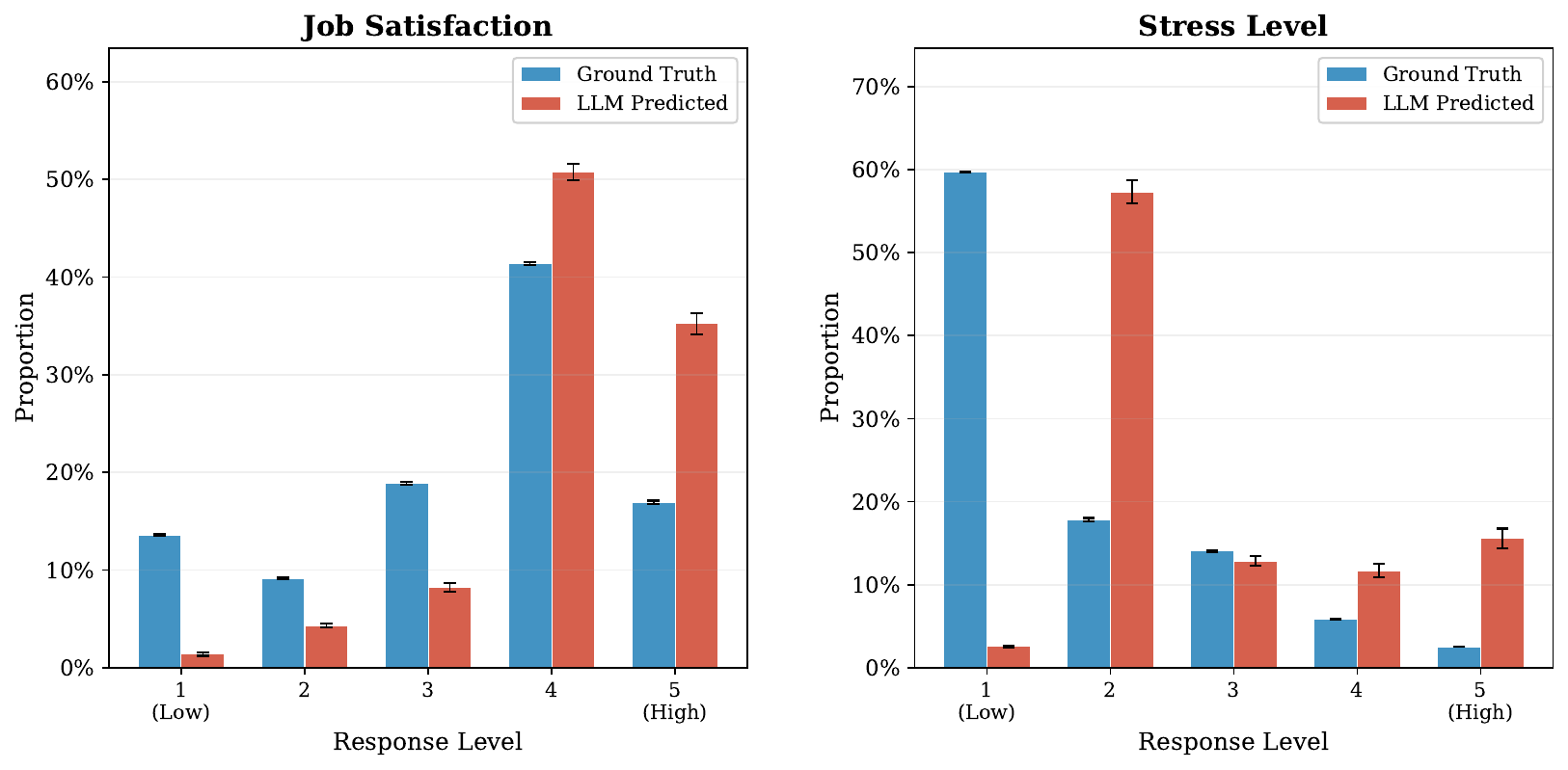}
\caption{\hl{Distributional comparison between ground-truth and LLM-predicted responses for JobSatisfaction and Stress (EmpS, LLaMA 3.1-8B, averaged over 3 runs). Both attributes exhibit skewed empirical distributions: JobSatisfaction is right-skewed (concentrated at high levels), whereas Stress is left-skewed (concentrated at low levels). The model successfully reproduces the skewed distribution for JobSatisfaction but fails for Stress, instead generating a more uniform distribution that inverts the dominant ground-truth category.}}
\label{fig:bias_amplification}
\end{figure}

\hl{Figure~\ref{fig:bias_amplification} compares the ground-truth and LLM-predicted response distributions for JobSatisfaction and Stress. Both attributes exhibit skewed empirical distributions: JobSatisfaction is right-skewed (concentrated at high satisfaction levels), whereas Stress is left-skewed (concentrated at low stress levels). Despite this structural similarity in the ground truth, the model's performance diverges markedly between the two attributes. For JobSatisfaction, the predicted distribution closely tracks the empirical baseline, capturing the right-skewed shape with mass concentrated at levels 4 and 5. This distributional alignment confirms that JobSatisfaction maintains a detectable statistical association with the available demographic profile, making it a suitable target for PAS. For Stress, however, the model fails to reproduce the heavily left-skewed distribution, which is predominantly concentrated at level~1 ($59.7\%$). Instead, the predicted distribution spreads mass more uniformly across levels 2 through 5, effectively inverting the dominant ground-truth category. This failure mode reveals that Stress levels are largely orthogonal to the demographic covariates available in EmpS and that the model cannot reliably infer this attribute from the provided profile features. The contrast between these two attributes underscores a fundamental insight for PAS deployment: the suitability of a target attribute for demographic-profile-based simulation depends critically on whether that attribute maintains a robust statistical association with the available covariates. Attributes like JobSatisfaction, which exhibit such associations, can be predicted with reasonable fidelity at both the individual and population levels. Conversely, attributes like Stress, which lack these associations, should be treated with caution in any downstream analysis relying on distributional fidelity.}

\section{Full Attribute Simulation (FAS)}
\label{sec:section4}
\hl{While PAS assumes that individual demographic records are available, a different and more demanding scenario arises when an organization possesses only aggregate population statistics rather than individual-level data. Consider a market research firm that has access to published census summaries for a target region, such as the joint distribution of age, gender, and income brackets, but holds no individual survey records. The firm needs to generate a synthetic population of respondents whose collective attribute distribution matches the known aggregate statistics, so that downstream analyses such as consumer preference estimation or policy impact assessment can be conducted. This is the setting that FAS is designed to address: the LLM is given population-level priors ({\it e.g.}, ``43\% of respondents are aged 25--44, median household income is \$62,000'') and asked to generate a set of synthetic individuals whose joint attribute distribution approximates these priors.} Large-scale sociological analysis frequently necessitates understanding population-level trends and aggregate distributions beyond individual-level responses, and FAS tackles precisely this challenge: generating comprehensive synthetic datasets that accurately preserve demographic distributions while inferring unobserved attributes.
As illustrated in Figure~\ref{fig:framework}, this capability is critical for policy formulation when only aggregated statistics ({\it e.g.}, age/gender distributions) are available, but granular, domain-specific insights ({\it e.g.}, opinions on urban development) remain unobserved.
\hl{Unlike PAS, which conditions on individual records and predicts missing attributes per person, FAS requires only aggregate statistics and generates an entirely synthetic population from scratch. It is therefore most appropriate as a diagnostic, methodological, or exploratory tool, for example to test analytical pipelines, generate synthetic training data, or explore hypothetical population scenarios, rather than as a substitute for real survey data collection~\cite{bisbee2024synthetic}.}

\subsection{Formulation}
The Full Attribute Simulation (FAS) formalizes the generation of multi-attribute survey samples through two complementary paradigms, each addressing distinct methodological requirements. 
Let $\mathcal{D}_{\text{syn}} = \{\mathbf{x}_1, \mathbf{x}_2, \dots, \mathbf{x}_N\}$ denote the synthetic dataset containing $N$ instances, where each $\mathbf{x}_i \in \mathbb{R}^M$ comprises $M$ attributes spanning demographic, behavioral, and contextual dimensions. 
FAS aims to approximate the joint distribution $P(\mathbf{X}_1, \mathbf{X}_2, \dots, \mathbf{X}_M)$ adhering to predefined constraints, implemented via two operational modes: \textbf{zero-context} and \textbf{context-enhanced} generation.

\paragraph{Zero-Context Generation:} 
The zero-context generation operates under the assumption that LLMs can probabilistically reconstruct plausible multivariate distributions solely from their parametric knowledge~\cite{cao-etal-2025-specializing,anthis2025llm}. 
Formally, the synthetic instances are sampled as:
\begin{equation}
\mathbf{x} \sim P_\theta(\mathbf{X} | \mathbf{C}, \mathbf{S}),
\end{equation}
where $\mathbf{C}$ encapsulates contextual metadata ({\it e.g.}, U.S. households in 2022), $\mathbf{S}$ defines schema constraints ({\it e.g.}, attribute types, value ranges), and $\theta$ represents the LLM’s parameters encoding implicit sociodemographic patterns. 
This paradigm eliminates external distributional guidance, relying entirely on the model’s emergent understanding of attribute correlations. 
For instance, when simulating energy consumption (RECS dataset), the schema specifies permissible ranges for \textit{annual kWh usage} ($500$--$15,000$) and \textit{employment status} (categorical options: employed/retired), while the LLM autonomously infers relationships between income brackets and electricity expenditure.

\paragraph{Context-Enhanced Generation:} 
In contrast, context-enhanced generation explicitly conditions the synthesis process on empirical priors, such as population-level statistics or attribute distributions, to better align synthetic generations with real-world patterns. Synthetic instances are sampled as:
\begin{equation}
\mathbf{x} \sim P_\theta(\mathbf{X} | \mathbf{C}, \mathbf{S}, P(\mathbf{A}_{\text{prior}})),
\end{equation}
where $P(\mathbf{A}_{\text{prior}})$ injects known marginal or conditional distributions ({\it e.g.}, age-gender cohorts by state, education-income correlations). 
This integration occurs through structured prompting. 
For example, appending U.S. Census-derived age distributions ($\mu=45.2$, $\sigma=18.3$) and party affiliation rates (Democrat: $43\%$, Republican: $37\%$) when simulating political surveys (ANES). LLMs then perform constrained sampling, iteratively refining proposals to align with provided prior distributions, akin to a learned rejection sampling algorithm.

To validate the performance of LLMs in the FAS scenario under the two generation modes, we select four representative datasets, maintaining the same LLMs and experimental conditions used in the PAS scenario. 
The evaluation is performed using the KL-based score (see Appendix~\ref{appendix:numerical_pred_task}), which assesses the alignment between the generated and ground-truth distributions. 

\begin{table}[htbp]
  \centering
  \small
  \setlength{\tabcolsep}{3pt}  
  \renewcommand{\arraystretch}{1.25}  
  \caption{LLM performance in FAS. The KL-based scores (↑) are reported as {\it avg.~$\pm$~std.}: bold for the best and underlined for the second-best. Color percentage indicates relative performance change from zero-context to context-enhanced generation: green for improvement and red for degradation.}
  \label{tab:fas_perf}
  \resizebox{\textwidth}{!}{
  \begin{tabular}{l|cccc|cccc}
    \Xhline{1.2pt}
    \multirow{2}{*}{\textbf{Dataset}} & \multicolumn{4}{c|}{\textbf{Zero-Context Generation}} & \multicolumn{4}{c}{\textbf{Context-Enhanced Generation}} \\
    \cline{2-5} \cline{6-9}
    & GPT-3.5 Turbo & GPT-4 Turbo & LLaMA 3.0 & LLaMA 3.1 & GPT-3.5 Turbo & GPT-4 Turbo & LLaMA 3.0 & LLaMA 3.1 \\
    \hline
    GSS & \uline{0.6490} & \textbf{0.6490} & 0.6508 & 0.6520
        & 0.6569 \calimp{1.22} & \textbf{0.6862} \calimp{5.73} & 0.6571 \calimp{0.98} & \uline{0.6726} \calimp{3.16} \\
    Trell SMU & \uline{0.3170} & \textbf{0.3320} & 0.2353 & 0.2992
              & \uline{0.3410} \calimp{7.57} & \textbf{0.3617} \calimp{8.95} & 0.2557 \calimp{8.68} & 0.2792 \calimp{-6.67} \\
    YPS & \uline{0.6674} & \textbf{0.7008} & 0.6567 & 0.6549
        & \uline{0.6764} \calimp{1.35} & \textbf{0.7210} \calimp{2.89} & 0.6569 \calimp{0.04} & 0.6539 \calimp{-0.16} \\
    RECS & \uline{0.6334} & \textbf{0.6384} & 0.5597 & 0.3826
         & \uline{0.6092} \calimp{-3.81} & \textbf{0.6403} \calimp{0.30} & 0.5883 \calimp{5.12} & 0.5529 \calimp{44.51} \\
    \Xhline{1.2pt}
  \end{tabular}
  }
\end{table}

\subsection{FAS Performance Analysis} 
Table~\ref{tab:fas_perf} presents the FAS performance of various LLMs, providing valuable insights into their ability to adapt to different generation paradigms, especially when additional context is provided. 
In general, GPT-4 Turbo demonstrates superior performance across all datasets and generation paradigms. 
This overall superiority is likely due to the larger parameter scale of GPT models, which enables them to better handle complex reasoning and inference tasks. 
The increased scale allows GPT-4 Turbo to process and integrate additional context more effectively than the 8B versions of LLaMA models selected here, contributing to its superior performance in the FAS scenario. 
In addition, we observe that the top-performing and second-best LLMs (marked with bold and underlined) generally belong to the same LLM series, showing consistency across datasets.

On the other hand, the context-enhanced generation generally achieves better results than the zero-context generation, with most models showing performance improvements when given additional context. 
Notably, GPT-4 Turbo demonstrates greater contextual utilization than its GPT-3.5 counterpart, achieving larger performance gains across datasets. 
This implies that integrated context helps models refine decisions through latent knowledge activation for the FAS scenario. 
Nevertheless, some models exhibit slight performance declines ({\it e.g.}, GPT-3.5 Turbo on RECS; LLaMA 3.1 on Trell SMU and YPS). 
These exceptions indicate that context effectiveness remains dataset- and model-dependent, with variations in contextual processing capabilities across architectures. 

\begin{figure}[htbp]
\centering
\includegraphics[width=0.7\textwidth]{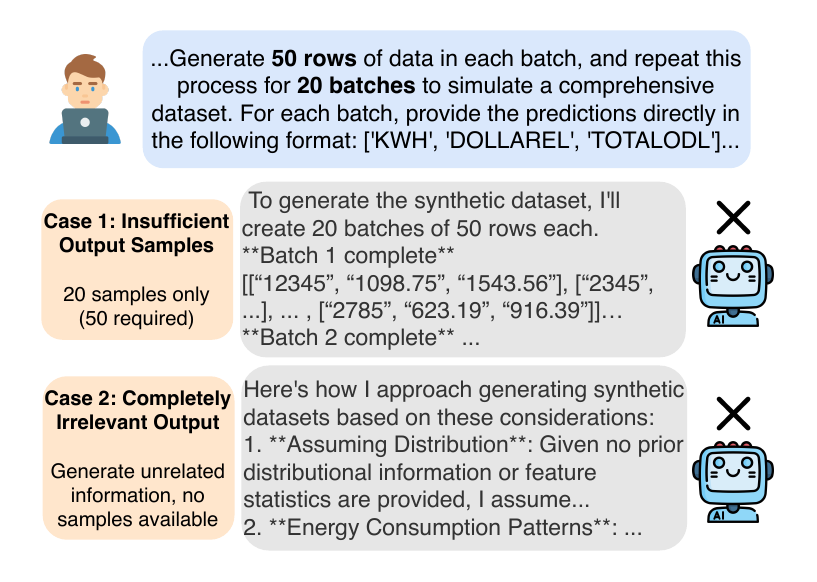}
\caption{Representative error cases in FAS.}
\label{fig:fas_failure_error_cases}
\end{figure}

\subsection{FAS Failure Phenomenon}
\label{sec:fas_failure}
A major challenge in the FAS scenario is the occurrence of FAS failures, where a significant proportion of generated outputs do not meet our requirements, leading to a loss of usable samples. 
We identify the following two failure modes, with representative examples shown in Figure~\ref{fig:fas_failure_error_cases}:

\begin{itemize}
\item \textbf{Case 1: Insufficient Output Sample}  

In certain cases, the model fails to produce the required number of samples, leading to an incomplete representation of the underlying statistical distribution.
\item \textbf{Case 2: Completely Irrelevant Output}  

The model generates non-structured outputs, such as textual descriptions, instead of the expected structured data table, making it unsuitable for quantitative analysis. 
\end{itemize}

\begin{figure}[htbp]
\centering
\includegraphics[width=0.7\textwidth]{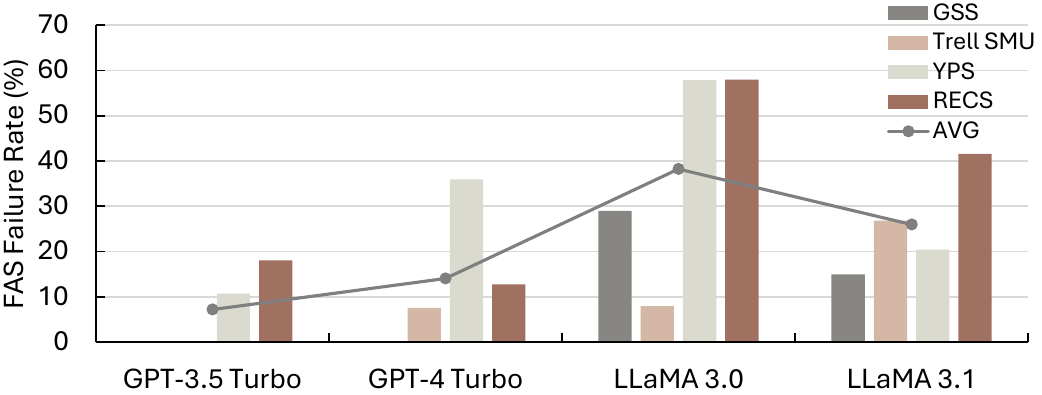}
\caption{Statistical analysis of failure rates across different attributes in FAS.}
\label{fig:fas_failure_rate}
\end{figure}

Experimental results in Figure~\ref{fig:fas_failure_rate} reveal significant performance disparities: LLaMA variants exhibit $15\text{--}58\%$ failure rates across all datasets. In contrast, GPT models maintain near-zero failure rates on simpler tasks ({\it e.g.}, GSS and Trell SMU), but show moderate errors on more complex datasets. 
This divergence principally originates from two architectural advantages in the GPT series. 
(i) \textbf{Model Size}: their superior parametric capacity (175B {\it vs.} 8B) enables precise modeling of complex conditional distributions, which is a critical requirement for attribute simulation \cite{gpt3,kaplan2020scaling}. 
(ii) \textbf{Alignment Optimization}: GPT models benefit from advanced alignment strategies such as Reinforcement Learning from Human Feedback (RLHF) \cite{gpt4tech,rlhf}, which enhance their ability to follow instructions and maintain structured output fidelity.
In contrast, LLaMA 3 employs a lighter alignment pipeline comprising Supervised Fine-Tuning (SFT), Direct Preference Optimization (DPO), and rejection sampling \cite{llama3, dpo}, which improves efficiency and training scalability but may underperform on tasks requiring strict format adherence \cite{yao2025reff}.

\section{Discussions}
\hl{The baseline experiments establish the capabilities and failure modes of LLMs for survey simulation tasks. In this section, we investigate three optimization dimensions that directly affect simulation fidelity: (1) the role of reasoning-oriented model distillation in enhancing attribute inference, (2) the impact of contextual prompting on improving response accuracy, and (3) the sensitivity of few-shot evaluation to example balance. Together, these analyses provide actionable guidelines for deploying LLMs in sociological research while surfacing remaining challenges.}

\hl{It is important to emphasize that our evaluation focuses on statistical and distributional similarity rather than behavioral fidelity or construct validity. This distinction has direct implications for how the findings below should be interpreted: improvements in accuracy or KL-based scores reflect better alignment with empirical distributions, not necessarily a deeper understanding of the social processes that generate those distributions.}

\subsection{Leveraging Reasoning-Oriented Distilled Models}
\begin{figure}[htbp]
\centering
\includegraphics[width=0.7\textwidth]{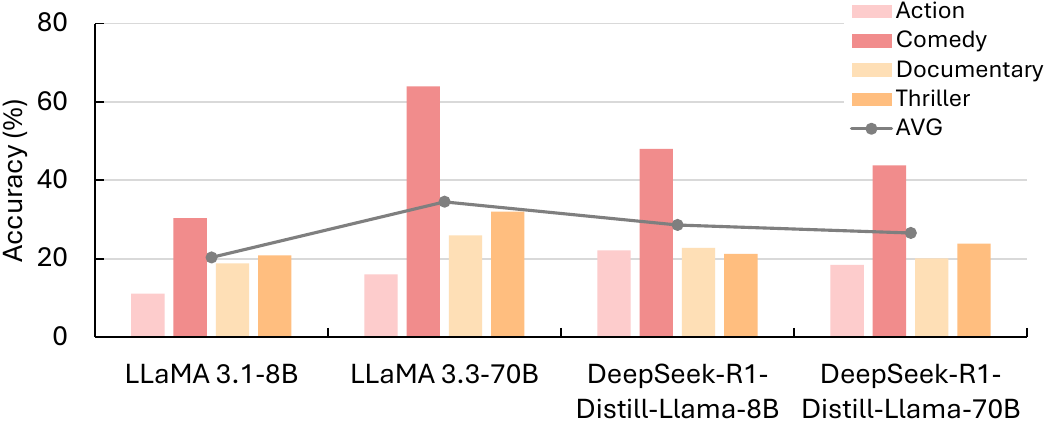}
\caption{Performance Comparison between LLaMA models and their DeepSeek-R1-Distill counterparts.}
\label{fig:perf_inference_llm}
\end{figure}

\hl{Inferring survey attributes from incomplete respondent profiles requires nontrivial reasoning over attribute interdependencies. This section investigates whether distilled LLMs explicitly optimized for reasoning, such as those from the DeepSeek-R1-Distill family~\cite{deepseek}, offer advantages over standard instruction-tuned models in survey simulation settings.}
To evaluate their effectiveness, we compare sociodemographic response generation on the YPS dataset between standard LLaMA models (3.1-8B and 3.3-70B) and their reasoning-optimized distilled counterparts (DeepSeek-R1-Distill-Llama-8B and -70B).
Figure~\ref{fig:perf_inference_llm} reveals that the reasoning-oriented distillation leads to a notable performance improvement for the 8B LLaMA model, yielding a relative increase of 40.6\% and indicating enhanced response generation capabilities. However, for the 70B LLaMA model, the distillation does not provide a significant advantage, instead resulting in a relative decrease of 22.9\%.

These findings suggest that while the reasoning-oriented distillation effectively boosts the performance of smaller models, its benefits diminish in larger models.
One possible explanation is that larger models, with their extensive parameter space, already possess the capacity to generate high-quality responses without requiring distillation \cite{kaplan2020scaling}, and applying further reasoning-oriented distillation may induce excessive internal deliberation in complex tasks, potentially impairing output accuracy or coherence \cite{cuadron2025danger}.
In contrast, smaller models, limited by fewer parameters, benefit more from the distillation as it helps streamline their processing and enhances their ability to simulate accurate sociodemographic responses \cite{ramesh2025generalization}.
\hl{These results suggest that practitioners working with resource-constrained deployments can achieve meaningful accuracy gains by substituting standard small models with their reasoning-distilled counterparts, whereas large-model deployments should be evaluated on a case-by-case basis before adopting distillation.}

\subsection{Context-Aware Prediction Enhancement}
\hl{Survey responses are inherently situated within specific sociopolitical, temporal, and cultural contexts. A respondent’s attitudes toward environmental policy, for example, cannot be fully understood without reference to the prevailing policy landscape and recent events. This section investigates whether explicitly contextualizing prompts with domain-relevant information improves the fidelity of LLM-generated survey responses.}
To demonstrate the impact of integrating prompt-based contextual information, we show that embedding precise contextual cues, such as explicit references to domains, timeframes, or real-world policy frameworks, significantly enhances prediction accuracy in sociodemographic response generation.
Our experiments with the LLaMA 3.1-8B model reveal that the contextual information enables the model to activate latent knowledge and produce predictions that are more closely aligned with empirical scenarios.
To evaluate the effectiveness of context-aware prediction, we concentrated on four key domains of public opinion concerning government spending in the GSS dataset. Contextual information was introduced to guide the model’s response generation. For example, the prompt ``{\it Considering the environmental policies of the United States in 2022}’’ was used to provide contextual grounding in the environmental domain.

\begin{figure}[htbp]
\centering
\includegraphics[width=0.7\textwidth]{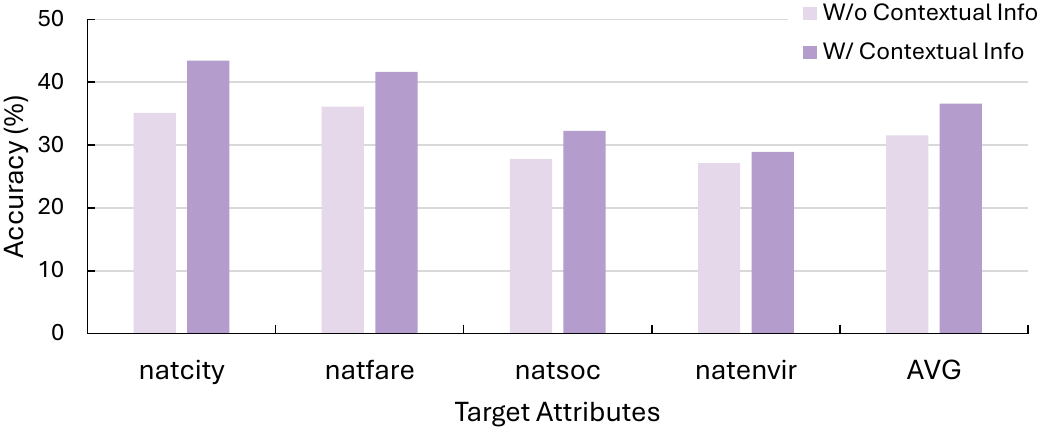}
\caption{Impact of contextual prompts on predictive accuracy in multiple-choice tasks.
}
\label{fig:context_inte}
\end{figure}

Incorporating contextual prompts led to improved prediction performance across the four domains, namely environmental protection (\textit{natenvir}), big city issues (\textit{natcity}), welfare programs (\textit{natfare}), and social security (\textit{natsoc}), with relative gains ranging from 6.67\% to 23.62\% (see Figure~\ref{fig:context_inte}).
These results establish context integration as a critical requirement for survey simulation systems, governed by two synergistic mechanisms: \textbf{first}, contextual anchoring overrides LLMs’ tendency for shallow pattern matching when addressing complex societal issues;
\textbf{second}, domain-specific knowledge activation closes the abstraction gap between model outputs and tangible policy dynamics~\cite{Sahoo2024}.
\hl{These findings suggest that survey simulation practitioners should embed domain-specific contextual cues in prompts whenever the target survey domain is known. The consistent performance gains observed across diverse policy domains indicate that context-aware prompting is a broadly applicable strategy for improving simulation fidelity, regardless of the specific task setting.}
This shift ensures that LLMs evolve from mere statistical pattern extractors into contextually grounded reasoning systems capable of supporting societal decision modeling.

\subsection{Mitigating In-Context Learning Bias}
\label{sec:few_shot_bias}
\hl{Few-shot learning enables LLMs to adapt to specific tasks with limited examples, but the selection and balance of these examples can systematically bias model outputs. In survey simulation, where demographic representation and response distributions carry substantive meaning, such biases pose significant validity threats. This section examines how example distribution affects prediction accuracy in few-shot survey simulation settings.}

Prior research demonstrates that few-shot learning exhibits predictive bias under class-imbalanced example distributions \cite{few_shot_class_imbalance}, advocating balanced sampling to mitigate bias and enhance modeling validity.
Building on this insight, we examine our setting focused on respondents’ travel mode on travel days (i.e., \textit{TRPTRANS}) in the NHTS dataset, and observe significant accuracy fluctuations driven by different example distributions.

\begin{figure}[htbp]
    \centering
    \includegraphics[width=0.7\textwidth]{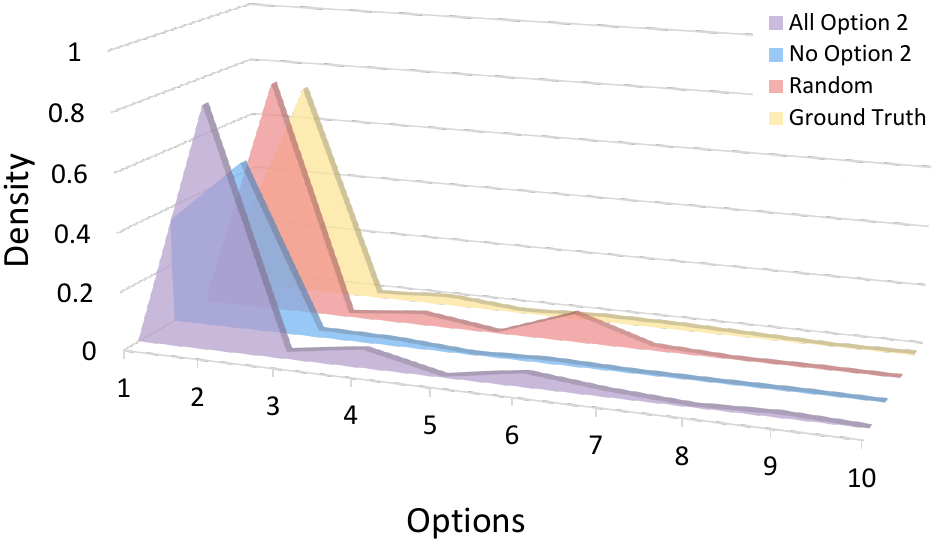}
    \caption{Comparison of output distributions under different few-shot example configurations.}
    \label{fig:imbalance}
\end{figure}

We manipulated example distributions using three configurations and analyzed their output distributions, highlighting the ratio of option 2: (i) \textbf{all option 2} (82.38\%), (ii) \textbf{no option 2} (57.76\%),  and (iii) \textbf{random selection} (80.01\%).
Notably, configuration (ii) caused a 24.62 percentage point decline compared to (i) (see Figure~\ref{fig:imbalance}), underscoring the strong influence of embedded input distributions on model outputs.
Statistical evidence from our experiments reveals a strong dependency between example balance and predictive robustness in our few-shot evaluation, directly motivating our adoption of strict example balancing measures.
\hl{These findings suggest that practitioners employing few-shot learning for survey simulation should draw demonstrations from balanced class distributions and treat the sensitivity of model outputs to example composition as a systematic evaluation criterion rather than an implementation detail.}

\subsection{Ethical Considerations}
\hl{The use of LLMs to generate synthetic survey responses raises four ethical considerations that must be addressed before these technologies can be widely adopted in sociological research:}

\hl{(1) \textbf{Bias amplification.} Our experiments demonstrate that LLMs can amplify existing biases present in training data. For example, predictions about income based on education level may exhibit more pronounced gender or racial disparities than those observed in ground-truth data. This risk necessitates rigorous bias audits before deploying LLM-based survey simulation tools.}

\hl{(2) \textbf{Representation and generalizability.} The datasets evaluated in this study are predominantly drawn from North American and European contexts, reflecting the current availability of large-scale public survey data rather than a fundamental constraint of the evaluation framework. The benchmark is designed to be extensible: as cross-cultural and multilingual survey datasets become available, the same evaluation protocol can be applied to assess LLM performance in diverse sociological contexts.}

\hl{(3) \textbf{Risk of misuse or misrepresentation.} Without proper disclosure, LLM-generated responses could be mistaken for genuine human survey data, potentially leading to flawed research conclusions or policy decisions. Practitioners should explicitly disclose when synthetic data has been used and clarify the limitations of LLM-generated responses.}

\hl{(4) \textbf{Transparency and reproducibility.} Synthetic data generation processes should be transparent and reproducible. In contexts with limited data availability, synthetic data should be used as a complement to, rather than a replacement for, genuine survey data.}

\subsection{Limitations and Future Research Directions}
\hl{Despite the strengths of our framework and benchmark, our work has several limitations that suggest directions for future research:}
\hl{(1) Our evaluation focuses on statistical and distributional similarity rather than behavioral fidelity or construct validity, which are important considerations in social science research;}
\hl{(2) The datasets evaluated in this study are predominantly drawn from North American and European contexts; extending the benchmark to cross-cultural and multilingual settings is an important direction for future work;}
\hl{(3) While we have examined partial and full attribute simulation as complementary approaches, further research is needed to explore hybrid methods that combine elements of both;}
\hl{(4) Our analysis of failure cases, while significant, is not exhaustive, and future work should continue to identify and address the limitations of LLM-based survey simulation.}

\section{Conclusion}
Questionnaire-based surveys remain indispensable for understanding public sentiment and informing evidence-based policy, yet their high cost, lengthy timelines, and declining response rates motivate the exploration of LLM-based simulation as a complementary analytical tool.
\hl{To support systematic evaluation of this emerging paradigm, this work formalizes two task settings that reflect distinct real-world data availability conditions: Partial Attribute Simulation (PAS), where individual demographic records are available but attitudinal attributes are missing, and Full Attribute Simulation (FAS), where only aggregate population statistics are available and a synthetic population must be generated from scratch. A unified benchmark, \textbf{LLM-S\textsuperscript{3}}, spanning eleven real-world datasets across four sociological domains, is curated to enable consistent, reproducible evaluation across models and settings.}

\hl{Experiments across four representative LLMs yield several actionable insights. In PAS, the benchmark reveals which survey constructs maintain detectable statistical associations with available demographic profiles, enabling practitioners to identify suitable simulation targets before deployment. In FAS, the evaluation surfaces how model scale and alignment strategy jointly shape the fidelity of synthetic population generation. Taken together, these findings position LLM-S\textsuperscript{3} as a diagnostic instrument that helps researchers calibrate expectations, select appropriate models, and design more effective prompting strategies for LLM-assisted survey research. We hope this benchmark serves as a foundation for future work on broadening the sociological scope, improving cross-cultural coverage, and advancing the methodological rigor of LLM-based survey simulation.}

\bibliographystyle{ACM-Reference-Format}
\bibliography{main}

\appendix

\newpage
\appendix

\section{Experimental Details}
\label{sec:appendix_A}

\subsection{Details of Survey Datasets from Different Domains}
In this study, we undertake a comprehensive evaluation of large language models (LLMs) by leveraging a diverse array of public survey datasets across four distinct domains: \textbf{Social and Public Affairs}, \textbf{Work and Income}, \textbf{Household and Behavioral Patterns}, and \textbf{Health and Lifestyle}. By utilizing datasets such as the American National Election Studies (ANES), General Social Survey (GSS), and the Residential Energy Consumption Survey (RECS), we ensure a multifaceted approach that captures a wide range of sociocultural factors and individual experiences. This extensive selection of datasets not only enhances the robustness of our benchmark but also allows for a nuanced understanding of how LLMs can simulate responses reflective of varied demographic backgrounds and social contexts. Through this work, we aim to contribute valuable insights into the capabilities of LLMs as virtual survey respondents, highlighting their potential in generating contextually relevant and representative sociological data. Table~\ref{tab:dataset-desciption} lists details of each dataset.
\begin{table}[htbp]
    \centering
    \scriptsize
    \setlength{\tabcolsep}{3pt}  
    \renewcommand{\arraystretch}{1.5} 
    \caption{Summary of the survey datasets utilized in constructing the benchmark for this study. The numbers of cases and attributes represent the total sample size and variable count in each raw dataset, respectively.}
    \label{tab:dataset-desciption}
    \resizebox{\textwidth}{!}{
    \begin{tabular}{> {\centering\arraybackslash}m{2.5cm} | >{\raggedright\arraybackslash}m{3cm} | >{\centering\arraybackslash}m{1cm} | > {\centering\arraybackslash}m{1.5cm} | >{\raggedright\arraybackslash}m{4cm} | >{\centering\arraybackslash}m{1.9cm} | > {\centering\arraybackslash}m{1cm}}
        \Xhline{1.2pt}
        \rowcolor{gray!10}
        \textbf{Domain} & \textbf{Survey Name} & \textbf{Abbr.} & \textbf{Year} & \textbf{Description} & \textbf{\# Cases} & \textbf{\# Attrs.} \\
        \hline 
        \multirow{4.5}{=}{Social and Public Affairs } 
        & American National Election Studies & ANES & 2020 & Individual’s views on certain social and political issues. & 3080 individuals & 470 \\
        \cline{2-7}
        & General Social Survey & GSS & 2022 & Individual's attitudes towards work and elections. & 4150 individuals & 1156 \\
        \cline{2-7}
        & Basic Income Survey & BIS & 2016-2017 spring & A survey about Europeans' opinions toward basic income. & 9649 individuals & 15 \\
        \hline 

        \multirow{3.3}{=}{Work and Income} 
        & Employee Survey & EmpS & Not specified & A survey about employee life experience and job satisfaction. & 3025 individuals & 23 \\
        \cline{2-7}
        & American Community Survey & ACS & Not specified & The ACS dataset offers a nuanced exploration of socioeconomic dynamics in the United States. & 2000 individuals & 13 \\
        \hline 

        \multirow{4.5}{=}{Household and Behavioral Patterns} 
        & Residential Energy Consumption Survey & RECS & 2020 & Household survey, energy supplier data, and consumption/expenditure estimates. & 18496 households & 799 \\
        \cline{2-7}
        & National Household Travel Survey & NHTS & 2017 & Household’s mobility choices and daily travel behavior. & 923572 queries & 229 \\
        \cline{2-7}
        & Trell Social Media Usage Data & Trell SMU & 2020 & Social media user's viewing/uploading/commenting habits towards a certain video website. & 488877 individuals & 26 \\
        \hline 

        \multirow{4.5}{=}{Health and Lifestyle} 
        & Music \& Mental Health Survey Results & MxMH & 2022 & Individual’s music preferences and mental health correlations. & 736 individuals & 33 \\
        \cline{2-7}
        & Young People Survey & YPS & 2013 & Individual’s preferences, habits, and personality traits. & 1011 individuals & 150 \\
        \cline{2-7}
        & Mental Health Dataset & MHD & 2014-2016 & Survey results on music taste and self-reported mental health. & 292364 individuals & 17 \\     
        \Xhline{1.2pt}
    \end{tabular}
    }
\end{table}

\subsection{Evaluation Metrics}
\label{Metrics}
To assess the performance of the models across different tasks, we utilize distinct evaluation metrics tailored to the nature of each task. 

\subsubsection{Multiple-Choice Prediction Tasks}
\label{appendix:mc_pred_task}

For multiple-choice prediction tasks, each dataset comprises several discrete target attributes, each treated as a classification problem with a fixed number of answer options. 
Model performance score $S$ is evaluated using standard accuracy, defined for the \(j\)-th attribute as:
\begin{equation}
    S_j = \frac{1}{M_j} \sum_{i=1}^{M_j} \mathbb{I}(\hat{y}_{i}^{(j)} = y_{i}^{(j)}),
\end{equation}
where \(M_j\) is the number of instances for attribute \(j\), and \(\mathbb{I}(\cdot)\) is the indicator function. 
This metric quantifies the proportion of correct predictions.

\subsubsection{Numerical Prediction Tasks}
\label{appendix:numerical_pred_task}

For numerical prediction tasks, each dataset may contain multiple continuous target attributes (e.g., travel duration and distance on the travel days). The goal is to approximate the overall distribution of these attributes rather than predict individual values. 
To evaluate how well the model’s predicted distribution aligns with the ground truth, we first compute the Kullback–Leibler (KL) divergence for each target attribute:
\begin{equation}
    D_{\text{KL}}(P \parallel Q) = \sum_{i=1}^{K} P_i \log \frac{P_i}{Q_i},
\end{equation}
where \(P\) denotes the predicted distribution and \(Q\) denotes the ground-truth distribution. 
To improve interpretability and facilitate comparisons across models and datasets, we define a performance score $S$ as the normalized reciprocal of KL divergence: 
\begin{equation}
    S_j = \frac{\ln\left(1 + \frac{1}{D_j}\right)}{1 + \ln\left(1 + \frac{1}{D_j}\right)},
\end{equation}
which smoothly maps the divergence value \( D_j \in (0, +\infty) \) to a bounded interval \( (0, 1) \), where higher values indicate better alignment between the predicted and true distributions. 
Lower KL divergence values yield higher scores (better performance), whereas larger values result in scores closer to 0 (poorer performance). 
For both prediction tasks, we evaluate performance separately for each target attribute, and the final score reported in Tables~\ref{tab:pas_perf_numer} and~\ref{tab:pas_perf_mc} corresponds to the average performance $S_{\text{avg}}$ across all target attributes within each dataset: 
\begin{equation}
    S_{\text{avg}} = \frac{1}{N} \sum_{j=1}^{N} S_j,
\end{equation}
where \( N \) is the number of target attributes and $S_j$ denotes the performance score for the $j$-$th$ attribute. A higher average score indicates better overall performance across all attributes.  

\subsection{Attribute Configuration \& Detailed Performance}
Given the extensive number of attributes present in the original surveys, we collaborated with domain experts to identify a subset of representative and vital attributes that are crucial for our tasks. This selection process ensures that the chosen attributes not only encapsulate the complexity of the respondents' experiences but also enhance the relevance and effectiveness of the generated responses. For the predictive attributes, we focused on those that accurately reflect the respondents’ lifestyle habits, thereby providing deeper insights into their behaviors and preferences. 
By employing this targeted approach, our research aims to evaluate the performance of large language models in generating contextually appropriate and meaningful responses, ultimately advancing the methodology of virtual survey respondents across multiple sociocultural domains. Based on this attribute selection principle, we designed the following prediction tasks:

\textbf{PAS:} Table~\ref{tab:pas_attri} summarizes the attribute configuration used for prediction tasks in the PAS scenario, preserving the original survey attribute names. 
Attributes in the ``Input Attributes'' column that appear with underscores are withheld when serving as prediction targets. 
Italicized entries indicate grouped attribute sets not shown in full. 
Tables~\ref{tab:pas_perf_mc_detail} (multiple-choice) and \ref{tab:pas_perf_numer_detail} (numerical) report the corresponding results, with performance (↑) presented as mean~$\pm$~std. over repeated runs and bold highlighting the best result.

\begin{table}[htbp]
  \centering
  \tiny
  \setlength{\tabcolsep}{2pt}
  \renewcommand{\arraystretch}{2}
  \caption{Attribute selection for prediction tasks in the PAS scenario. 
  A subset of variables was chosen as target attributes based on domain expertise, with input attributes selected to ensure scientific rigor, and original survey attribute names retained. 
  Attributes marked with an underscore in the ``Input Attributes'' column are excluded when serving as targets, while italicized entries denote attribute sets not listed in detail. 
  Prediction tasks are classified as Multi-Choice or Numerical.}
  \label{tab:pas_attri}
  \resizebox{\textwidth}{!}{
  \begin{tabular}{l|
  >{\raggedright\arraybackslash}p{0.2\textwidth}|
  >{\raggedright\arraybackslash}p{0.37\textwidth}|
  >{\raggedright\arraybackslash}p{0.12\textwidth}|
  >{\raggedright\arraybackslash}p{0.12\textwidth}}
    \Xhline{1pt}
    \multirow{2}{*}{\textbf{Dataset}} 
    & \multirow{2}{*}{\textbf{Simulation Objectives}}
    & \multirow{2}{*}{\textbf{Input Attributes}}
    & \multicolumn{2}{c}{\textbf{Target Attributes}} \\
    \cline{4-5}
    & & & \textbf{Multi-Choice} & \textbf{Numerical} \\
    \hline
    ANES & Participation in political activities. The level of support for the candidate. & age, home\_ownership, income, vote20turnoutjob, particip\_count, \uline{meeting}, \uline{online}, \uline{moneyorg}, protest, persuade, button, ftmikepence1, ftrubio1, ftthomas1, ftfauci1, \uline{ftobama1}, \uline{fttrump1}, \uline{ftbiden1} & 2-Options: meeting, online, moneyorg & ftobama1, ftbiden1, fttrump1 \\
    \hline
    GSS & The attitude towards relieving work pressure. Public opinions. & age,  sex,  racecen1, sibs, marital, childs, region, educ, relig, income, wrkstat, occ10, hhtype1, hompop, polviews, natenvir, natheal, natcity, \uline{natcrime}, \uline{nateduc}, natarms, natfare, natsoc & 5-Options: strmgtsup, allorglevel & nateduc, natcrime \\
    \hline
    BIS & Vote for basic income. / Support basic income or not (reasons). & country\_code, age, gender, rural, dem\_education\_level, dem\_full\_time\_job, dem\_has\_children, question\_bbi\_2016wave4\_basicincome\_awareness / \uline{vote} / effect / \uline{argumentsfor} / argumentsagainst & 5-Options: vote & argumentsfor \\
    \hline
    RECS & Household fuel usage. & HHSEX, HHAGE, EMPLOYHH, state\_postal, EDUCATION, HOUSEHOLDER\_RACE, NHSLDMEM, ATHOME, MONEYPY, \uline{UGASHERE}, USEEL, USENG, USELP, USEFO, USESOLAR, USEWOOD, ALLELEC & 2-Options: UGASHERE & KWH \\
    \hline
    ACS & The working hours or income. & age, race, gender, birth\_qrtr, citizen, lang, edu, married, disability, employment, time\_to\_work & - & hrs\_work, income \\
    \hline
    Trell SMU & Video watching habit. / Total number of videos watched (normalized). & age\_group, gender, city-tier, avgCompletion, avgTimeSpent, avgDuration, avgComments, creations, \uline{content\_views}, num\_of\_comments, \uline{weekends\_trails\_watched\_per\_day}, \uline{weekdays\_trails\_watched\_per\_day} & 6-Options: \makecell{weekdays\_trails\_\\watched\_per\_day} & content\_views \\
    \hline
    EmpS & Levels of satisfaction with the job (5 levels). & \uline{WorkEnv}, EmpID, Age, Gender, MaritalStatu, JobLevel, Dept, EmpType, CommuteMode, EduLevel, haveOT, \uline{JobSatisfaction}, Experience, PhysicalActivityHours, SleepHours, CommuteDistance, NumCompanies, TeamSize, NumReports, EduLevel, TrainingHoursPerYear & 5-Options: WLB, WorkEnv, Workload, Stress, JobSatisfaction & - \\
    \hline
    NHTS & Trip model, trip duration, trip distance. & HHFAMINC, HOMEOWN, URBAN, HH\_CBSA, DRIVER, EDUC, R\_RACE, R\_SEX\_IMP, WORKER, TRAVDAY, \uline{TRPTRANS}, GASPRICE, TRAVDAY, TDAYDATE & 10-Options: TRPTRANS & TRPMILES, TRVLCMIN \\
    \hline
    MxMH & Mental health situation. & Timestamp, Age, Fav genre, Hours per day, While working, Instrumentalist, Composer, Exploratory, Classical, Country, EDM, Folk, Gospel, Hip hop, Jazz, K pop, Latin, Lofi, Metal, Pop, R\&B, Rap, Rock, Video game music & 11-Options: Anxiety, Depression, Insomnia, OCD & - \\
    \hline
    YPS & Degree of interests of movies and music. & Music, Slow songs or fast songs, Dance, Folk, Country, Classical music, Musical, Pop, Rock, Metal or Hardrock, Punk, Hiphop, Rap, Reggae, Ska, Swing, Jazz, Rock n roll, Alternative, Latino, Techno, Trance, Opera, Movies, Horror, Romantic, Sci-fi, War, Fantasy/Fairy tales, Animated & 5-Options: Action, Documentary, Thriller, Comedy& - \\
    \hline
    MHD & Feeling the level of repression. Coping\_Struggles only yes or no selection, others are three selections. & \textit{History of mental condition}, Timestamp, Gender, Country, Occupation, self\_employed, family\_history, treatment, Days\_Indoors,  Changes\_Habits, Mental\_Health\_History, mental\_health\_interview,  care\_options, Work\_Interest & Growing\_Stress, Mood\_Swings, Social\_Weakness & - \\
    \Xhline{1pt}
  \end{tabular}
  }
\end{table} 

\begin{table}[htbp]
  \centering
  \scriptsize
  \setlength{\tabcolsep}{2pt}
  \renewcommand{\arraystretch}{1.75}
  \caption{Performance of LLMs on multiple-choice prediction tasks in the PAS scenario. }
  \label{tab:pas_perf_mc_detail}
  \resizebox{\textwidth}{!}{
  \begin{tabular}{l|l|l|l|l|l|l|l|l|l}
    \Xhline{1pt}
    \multirow{2}{*}{\textbf{Dataset}} & \multirow{2}{*}{\textbf{Target Attributes}} &
    \multicolumn{4}{c|}{\textbf{Zero-Shot Evaluation}} &
    \multicolumn{4}{c}{\textbf{Few-Shot Evaluation}} \\
    \cline{3-10}
    & & GPT-3.5 Turbo & GPT-4 Turbo & LLaMA 3.0 & LLaMA 3.1 & GPT-3.5 Turbo & GPT-4 Turbo & LLaMA 3.0 & LLaMA 3.1 \\
    \hline
    \multirow{3}{*}{ANES}
      & meeting & \valuewithoutimp{0.7898}{0.0650} & \valuewithoutimp{\textbf{0.8440}}{0.0239} & \valuewithoutimp{0.8314}{0.0602} & \valuewithoutimp{0.7619}{0.0099}
      & \valuewithoutimp{0.8002}{0.0527} & \valuewithoutimp{\textbf{0.8353}}{0.0223} & \valuewithoutimp{0.7803}{0.0829} & \valuewithoutimp{0.7388}{0.0536} \\
      \cline{2-10}
      & online & \valuewithoutimp{0.7431}{0.0876} & \valuewithoutimp{\textbf{0.8069}}{0.0618} & \valuewithoutimp{0.7120}{0.0262} & \valuewithoutimp{0.6980}{0.0296}
      & \valuewithoutimp{0.7334}{0.0523} & \valuewithoutimp{\textbf{0.8059}}{0.0541} & \valuewithoutimp{0.7134}{0.0215} & \valuewithoutimp{0.6789}{0.0354} \\
      \cline{2-10}
      & moneyorg & \valuewithoutimp{0.8233}{0.0267} & \valuewithoutimp{\textbf{0.8566}}{0.0116} & \valuewithoutimp{0.8047}{0.0597} & \valuewithoutimp{0.7975}{0.0721}
      & \valuewithoutimp{0.8051}{0.0290} & \valuewithoutimp{\textbf{0.8244}}{0.0422} & \valuewithoutimp{0.8122}{0.0101} & \valuewithoutimp{0.7887}{0.0584} \\
    \hline
    \multirow{2}{*}{GSS}
      & strmgtsup & \valuewithoutimp{0.5225}{0.0777} & \valuewithoutimp{\textbf{0.5893}}{0.0064} & \valuewithoutimp{0.5563}{0.0826} & \valuewithoutimp{0.5793}{0.0064}
      & \valuewithoutimp{0.4625}{0.0356} & \valuewithoutimp{\textbf{0.6016}}{0.0264} & \valuewithoutimp{0.5289}{0.1266} & \valuewithoutimp{0.4868}{0.1217} \\
      \cline{2-10}
      & allorglevel & \valuewithoutimp{0.5950}{0.1940} & \valuewithoutimp{\textbf{0.6732}}{0.0615} & \valuewithoutimp{0.5016}{0.0339} & \valuewithoutimp{0.4145}{0.0588}
      & \valuewithoutimp{0.5270}{0.0503} & \valuewithoutimp{\textbf{0.6429}}{0.0433} & \valuewithoutimp{0.4577}{0.0523} & \valuewithoutimp{0.3389}{0.0338} \\
    \hline
    \multirow{1}{*}{BIS}
      & vote & \valuewithoutimp{0.1865}{0.0032} & \valuewithoutimp{0.2165}{0.0022} & \valuewithoutimp{0.2629}{0.0600} & \valuewithoutimp{\textbf{0.2848}}{0.0217}
      & \valuewithoutimp{0.1465}{0.0164} & \valuewithoutimp{0.2169}{0.0045} & \valuewithoutimp{\textbf{0.2261}}{0.0233} & \valuewithoutimp{0.2077}{0.0044} \\
    \hline
    \multirow{1}{*}{RECS}
      & UGASHERE & \valuewithoutimp{0.6208}{0.1116} & \valuewithoutimp{\textbf{0.8598}}{0.0793} & \valuewithoutimp{0.7257}{0.0144} & \valuewithoutimp{0.6540}{0.0730}
      & \valuewithoutimp{0.6386}{0.0257} & \valuewithoutimp{\textbf{0.9113}}{0.0165} & \valuewithoutimp{0.7483}{0.2132} & \valuewithoutimp{0.6742}{0.0623} \\
    \hline
    \multirow{1}{*}{Trell SMU}
      & weekdays\_trails\_... & \valuewithoutimp{0.6415}{0.0105} & \valuewithoutimp{\textbf{0.6525}}{0.0050} & \valuewithoutimp{0.3865}{0.0138} & \valuewithoutimp{0.5028}{0.0134}
      & \valuewithoutimp{0.7228}{0.0357} & \valuewithoutimp{0.6601}{0.0057} & \valuewithoutimp{0.7099}{0.0051} & \valuewithoutimp{\textbf{0.7309}}{0.0176} \\
    \hline
    \multirow{5}{*}{EmpS}
      & WLB & \valuewithoutimp{0.1849}{0.0114} & \valuewithoutimp{\textbf{0.1959}}{0.0016} & \valuewithoutimp{0.1951}{0.0228} & \valuewithoutimp{0.1804}{0.0027}
      & \valuewithoutimp{0.1728}{0.0073} & \valuewithoutimp{0.1913}{0.0070} & \valuewithoutimp{\textbf{0.1945}}{0.0147} & \valuewithoutimp{0.1898}{0.0026} \\
      \cline{2-10}
      & WorkEnv & \valuewithoutimp{0.1943}{0.0036} & \valuewithoutimp{0.1949}{0.0029} & \valuewithoutimp{\textbf{0.2080}}{0.0032} & \valuewithoutimp{0.1985}{0.0061}
      & \valuewithoutimp{0.1868}{0.0048} & \valuewithoutimp{0.1825}{0.0173} & \valuewithoutimp{0.2052}{0.0019} & \valuewithoutimp{\textbf{0.2071}}{0.0050} \\
      \cline{2-10}
      & Workload & \valuewithoutimp{0.2036}{0.0058} & \valuewithoutimp{0.1973}{0.0023} & \valuewithoutimp{0.2035}{0.0007} & \valuewithoutimp{\textbf{0.2136}}{0.0191}
      & \valuewithoutimp{0.1940}{0.0106} & \valuewithoutimp{0.1952}{0.0125} & \valuewithoutimp{\textbf{0.2001}}{0.0099} & \valuewithoutimp{0.1995}{0.0070} \\
      \cline{2-10}
      & Stress & \valuewithoutimp{0.1278}{0.0824} & \valuewithoutimp{0.1268}{0.0015} & \valuewithoutimp{\textbf{0.1847}}{0.0247} & \valuewithoutimp{0.1667}{0.0021}
      & \valuewithoutimp{0.1787}{0.0116} & \valuewithoutimp{0.1043}{0.0129} & \valuewithoutimp{0.1483}{0.0190} & \valuewithoutimp{\textbf{0.2425}}{0.0028} \\
      \cline{2-10}
      & JobSatisfaction & \valuewithoutimp{\textbf{0.3852}}{0.0434} & \valuewithoutimp{0.3706}{0.0004} & \valuewithoutimp{0.3598}{0.0779} & \valuewithoutimp{0.3348}{0.0614}
      & \valuewithoutimp{\textbf{0.4013}}{0.0118} & \valuewithoutimp{0.3848}{0.0029} & \valuewithoutimp{0.3252}{0.0696} & \valuewithoutimp{0.2327}{0.0049} \\
    \hline
    \multirow{1}{*}{NHTS}
      & TRPTRANS & \valuewithoutimp{\textbf{0.7314}}{0.0077} & \valuewithoutimp{0.7231}{0.0003} & \valuewithoutimp{0.5732}{0.0126} & \valuewithoutimp{0.6507}{0.0453}
      & \valuewithoutimp{0.7246}{0.0322} & \valuewithoutimp{\textbf{0.7300}}{0.0057} & \valuewithoutimp{0.6409}{0.0797} & \valuewithoutimp{0.5987}{0.0270} \\
    \hline
    \multirow{4}{*}{MxMH}
      & Anxiety & \valuewithoutimp{0.0650}{0.0036} & \valuewithoutimp{\textbf{0.0957}}{0.0000} & \valuewithoutimp{0.0704}{0.0083} & \valuewithoutimp{0.0898}{0.0175}
      & \valuewithoutimp{\textbf{0.1027}}{0.0332} & \valuewithoutimp{0.0888}{0.0065} & \valuewithoutimp{0.0695}{0.0104} & \valuewithoutimp{0.0771}{0.0105} \\
      \cline{2-10}
      & Depression & \valuewithoutimp{0.0966}{0.0414} & \valuewithoutimp{\textbf{0.1049}}{0.0010} & \valuewithoutimp{0.0892}{0.0144} & \valuewithoutimp{0.0972}{0.0167}
      & \valuewithoutimp{\textbf{0.0977}}{0.0169} & \valuewithoutimp{0.0889}{0.0158} & \valuewithoutimp{0.0963}{0.0067} & \valuewithoutimp{0.0970}{0.0175} \\
      \cline{2-10}
      & Insomnia & \valuewithoutimp{\textbf{0.1040}}{0.0702} & \valuewithoutimp{0.0903}{0.0010} & \valuewithoutimp{0.0828}{0.0113} & \valuewithoutimp{0.0876}{0.0123}
      & \valuewithoutimp{\textbf{0.1784}}{0.0618} & \valuewithoutimp{0.0968}{0.0056} & \valuewithoutimp{0.0897}{0.0090} & \valuewithoutimp{0.0709}{0.0080} \\
      \cline{2-10}
      & OCD & \valuewithoutimp{\textbf{0.2476}}{0.0367} & \valuewithoutimp{0.1174}{0.0052} & \valuewithoutimp{0.1179}{0.0177} & \valuewithoutimp{0.0986}{0.0137}
      & \valuewithoutimp{\textbf{0.2379}}{0.0948} & \valuewithoutimp{0.1165}{0.0091} & \valuewithoutimp{0.1125}{0.0150} & \valuewithoutimp{0.0992}{0.0120} \\
    \hline
    \multirow{4}{*}{YPS}
      & Action & \valuewithoutimp{\textbf{0.2388}}{0.0074} & \valuewithoutimp{0.2191}{0.0016} & \valuewithoutimp{0.2203}{0.0686} & \valuewithoutimp{0.2068}{0.0483}
      & \valuewithoutimp{0.2465}{0.0204} & \valuewithoutimp{\textbf{0.2939}}{0.0681} & \valuewithoutimp{0.1602}{0.0001} & \valuewithoutimp{0.1073}{0.0081} \\
      \cline{2-10}
      & Documentary & \valuewithoutimp{\textbf{0.2952}}{0.0026} & \valuewithoutimp{0.2698}{0.0018} & \valuewithoutimp{0.2731}{0.0157} & \valuewithoutimp{0.2885}{0.0232}
      & \valuewithoutimp{\textbf{0.2948}}{0.0119} & \valuewithoutimp{0.2758}{0.0303} & \valuewithoutimp{0.2809}{0.0186} & \valuewithoutimp{0.2092}{0.0823} \\
      \cline{2-10}
      & Thriller & \valuewithoutimp{0.2462}{0.0093} & \valuewithoutimp{\textbf{0.3482}}{0.0058} & \valuewithoutimp{0.3033}{0.0191} & \valuewithoutimp{0.2637}{0.0286}
      & \valuewithoutimp{\textbf{0.3214}}{0.0213} & \valuewithoutimp{0.3132}{0.0493} & \valuewithoutimp{0.2592}{0.0268} & \valuewithoutimp{0.2078}{0.0171} \\
      \cline{2-10}
      & Comedy & \valuewithoutimp{0.3115}{0.0104} & \valuewithoutimp{0.1865}{0.0059} & \valuewithoutimp{\textbf{0.4366}}{0.0880} & \valuewithoutimp{0.3557}{0.0702}
      & \valuewithoutimp{0.3035}{0.0597} & \valuewithoutimp{0.3044}{0.0405} & \valuewithoutimp{\textbf{0.4068}}{0.0243} & \valuewithoutimp{0.2716}{0.0596} \\
    \hline
    \multirow{3}{*}{MHD}
      & Growing\_Stress & \valuewithoutimp{0.3231}{0.0034} & \valuewithoutimp{0.3454}{0.0088} & \valuewithoutimp{\textbf{0.3461}}{0.0103} & \valuewithoutimp{0.3374}{0.0026}
      & \valuewithoutimp{0.3349}{0.0166} & \valuewithoutimp{0.3416}{0.0072} & \valuewithoutimp{\textbf{0.3420}}{0.0083} & \valuewithoutimp{0.3369}{0.0018} \\
      \cline{2-10}
      & Mood\_Swings & \valuewithoutimp{0.3446}{0.0050} & \valuewithoutimp{0.3225}{0.0026} & \valuewithoutimp{0.3362}{0.0037} & \valuewithoutimp{\textbf{0.3466}}{0.0078}
      & \valuewithoutimp{0.3322}{0.0161} & \valuewithoutimp{0.3185}{0.0103} & \valuewithoutimp{0.3428}{0.0114} & \valuewithoutimp{\textbf{0.3482}}{0.0062} \\
      \cline{2-10}
      & Social\_Weakness & \valuewithoutimp{0.3454}{0.0089} & \valuewithoutimp{0.3445}{0.0021} & \valuewithoutimp{0.3348}{0.0063} & \valuewithoutimp{\textbf{0.3461}}{0.0107}
      & \valuewithoutimp{0.3272}{0.0266} & \valuewithoutimp{0.3417}{0.0063} & \valuewithoutimp{0.3361}{0.0084} & \valuewithoutimp{\textbf{0.3460}}{0.0108} \\
    \Xhline{1pt}
  \end{tabular}
  }
\end{table} 
\begin{table}[htbp]
  \centering
  \scriptsize
  \setlength{\tabcolsep}{2pt}
  \renewcommand{\arraystretch}{1.75}
  \caption{Detailed Performance of LLMs on numerical prediction tasks in the PAS scenario. }
  \label{tab:pas_perf_numer_detail}
  \resizebox{\textwidth}{!}{
  \begin{tabular}{l|l|l|l|l|l|l|l|l|l}
    \Xhline{1pt}
    \multirow{2}{*}{\textbf{Dataset}} & \multirow{2}{*}{\textbf{Target Attributes}} &
    \multicolumn{4}{c|}{\textbf{Zero-Shot Evaluation}} &
    \multicolumn{4}{c}{\textbf{Few-Shot Evaluation}} \\
    \cline{3-10}
    & & GPT-3.5 Turbo & GPT-4 Turbo & LLaMA 3.0 & LLaMA 3.1 & GPT-3.5 Turbo & GPT-4 Turbo & LLaMA 3.0 & LLaMA 3.1 \\
    \hline
    \multirow{3}{*}{ANES}
      & ftobama1 & \valuewithoutimp{\textbf{0.5974}}{0.0218} & \valuewithoutimp{0.5888}{0.0163} & \valuewithoutimp{0.4745}{0.0813} & \valuewithoutimp{0.4660}{0.1026}
      & \valuewithoutimp{\textbf{0.6208}}{0.0008} & \valuewithoutimp{0.5499}{0.0754} & \valuewithoutimp{0.4631}{0.0949} & \valuewithoutimp{0.4823}{0.0822} \\
      \cline{2-10}
      & ftbiden1 & \valuewithoutimp{\textbf{0.5788}}{0.0376} & \valuewithoutimp{0.5378}{0.0540} & \valuewithoutimp{0.4909}{0.0666} & \valuewithoutimp{0.4823}{0.1231}
      & \valuewithoutimp{0.5666}{0.0148} & \valuewithoutimp{\textbf{0.5688}}{0.0583} & \valuewithoutimp{0.4654}{0.0724} & \valuewithoutimp{0.4529}{0.0173} \\
      \cline{2-10}
      & fttrump1 & \valuewithoutimp{0.5444}{0.0994} & \valuewithoutimp{\textbf{0.5461}}{0.0177} & \valuewithoutimp{0.4971}{0.0119} & \valuewithoutimp{0.4820}{0.0007}
      & \valuewithoutimp{0.4829}{0.0061} & \valuewithoutimp{\textbf{0.5191}}{0.0144} & \valuewithoutimp{0.4654}{0.0053} & \valuewithoutimp{0.4697}{0.0006} \\
    \hline
    \multirow{2}{*}{GSS}
      & nateduc & \valuewithoutimp{0.7062}{0.0138} & \valuewithoutimp{\textbf{0.7139}}{0.0041} & \valuewithoutimp{0.6983}{0.0221} & \valuewithoutimp{0.6798}{0.0089}
      & \valuewithoutimp{0.6988}{0.0060} & \valuewithoutimp{\textbf{0.7034}}{0.0276} & \valuewithoutimp{0.6959}{0.0118} & \valuewithoutimp{0.6753}{0.0366} \\
      \cline{2-10}
      & natcrime & \valuewithoutimp{0.6904}{0.0101} & \valuewithoutimp{\textbf{0.7024}}{0.0216} & \valuewithoutimp{0.6997}{0.0175} & \valuewithoutimp{0.6838}{0.0035}
      & \valuewithoutimp{0.6545}{0.0112} & \valuewithoutimp{\textbf{0.7073}}{0.0052} & \valuewithoutimp{0.6798}{0.0165} & \valuewithoutimp{0.6848}{0.0250} \\
    \hline
    \multirow{1}{*}{BIS}
      & argumentsfor & \valuewithoutimp{\textbf{0.6204}}{0.0462} & \valuewithoutimp{0.5119}{0.0129} & \valuewithoutimp{0.5777}{0.0458} & \valuewithoutimp{0.5730}{0.0378}
      & \valuewithoutimp{\textbf{0.6207}}{0.0417} & \valuewithoutimp{0.5299}{0.0337} & \valuewithoutimp{0.5640}{0.0486} & \valuewithoutimp{0.5622}{0.0476} \\
    \hline
    \multirow{1}{*}{RECS}
      & KWH & \valuewithoutimp{\textbf{0.6554}}{0.0162} & \valuewithoutimp{0.5648}{0.0568} & \valuewithoutimp{0.3114}{0.0801} & \valuewithoutimp{0.2669}{0.0197}
      & \valuewithoutimp{0.5293}{0.1865} & \valuewithoutimp{\textbf{0.5489}}{0.0920} & \valuewithoutimp{0.3088}{0.0874} & \valuewithoutimp{0.2365}{0.0226} \\
    \hline
    \multirow{2}{*}{ACS}
      & hrs\_work & \valuewithoutimp{0.5482}{0.1910} & \valuewithoutimp{0.4997}{0.2572} & \valuewithoutimp{\textbf{0.6461}}{0.1041} & \valuewithoutimp{0.5807}{0.1182}
      & \valuewithoutimp{0.5265}{0.1232} & \valuewithoutimp{0.3259}{0.0281} & \valuewithoutimp{\textbf{0.6506}}{0.1043} & \valuewithoutimp{0.6201}{0.0971} \\
      \cline{2-10}
      & income & \valuewithoutimp{0.4835}{0.0655} & \valuewithoutimp{0.4447}{0.1146} & \valuewithoutimp{0.5650}{0.0925} & \valuewithoutimp{\textbf{0.5725}}{0.1295}
      & \valuewithoutimp{0.4317}{0.0980} & \valuewithoutimp{0.4013}{0.0539} & \valuewithoutimp{0.5603}{0.1051} & \valuewithoutimp{\textbf{0.5738}}{0.1252} \\
    \hline
    \multirow{1}{*}{Trell SMU}
      & content\_views & \valuewithoutimp{0.3523}{0.0340} & \valuewithoutimp{0.4301}{0.0033} & \valuewithoutimp{\textbf{0.5612}}{0.0003} & \valuewithoutimp{0.5361}{0.0064}
      & \valuewithoutimp{0.4274}{0.0499} & \valuewithoutimp{0.5197}{0.0670} & \valuewithoutimp{\textbf{0.5462}}{0.0189} & \valuewithoutimp{0.4685}{0.0147} \\
    \hline
    \multirow{2}{*}{NHTS}
      & TRPMILES & \valuewithoutimp{0.5969}{0.0027} & \valuewithoutimp{0.4403}{0.1400} & \valuewithoutimp{0.5983}{0.1516} & \valuewithoutimp{\textbf{0.6851}}{0.0481}
      & \valuewithoutimp{0.5507}{0.1309} & \valuewithoutimp{0.4194}{0.2138} & \valuewithoutimp{\textbf{0.6435}}{0.1228} & \valuewithoutimp{0.5615}{0.0715} \\
      \cline{2-10}
      & TRVLCMIN & \valuewithoutimp{0.2050}{0.0104} & \valuewithoutimp{0.1847}{0.0283} & \valuewithoutimp{0.3871}{0.2022} & \valuewithoutimp{\textbf{0.5307}}{0.1590}
      & \valuewithoutimp{0.1629}{0.0684} & \valuewithoutimp{0.2011}{0.0860} & \valuewithoutimp{0.3381}{0.0887} & \valuewithoutimp{\textbf{0.5982}}{0.0805} \\
    \Xhline{1pt}
  \end{tabular}
  }
\end{table}

\textbf{FAS:} 
Table~\ref{tab:fas_attri} details the attribute selection for prediction tasks in the FAS scenario. 
Attributes marked with an underscore in the ``Input Attributes'' column are treated as optional, excluded in the zero-context generation, and included in the context-enhanced generation. 
Italicized entries refer to descriptive textual labels rather than specific attributes. 
Table~\ref{tab:fas_perf_detail} reports the performance for each target attribute, with bold values as the best.

\begin{table}[htbp]
  \scriptsize
  \setlength{\tabcolsep}{2pt}
  \renewcommand{\arraystretch}{2}
  \caption{Attribute selection details for prediction tasks in the FAS scenario. 
  Attributes with an underscore in the ``Input Attributes'' column are optional, excluded in the zero-context paradigm and included in the context-enhanced paradigm. 
  Italicized terms refer to textual descriptions rather than specific attributes.}
  \label{tab:fas_attri}
  \resizebox{\textwidth}{!}{
  \begin{tabular}{l|p{0.2\textwidth}|p{0.35\textwidth}|p{0.25\textwidth}}
    \Xhline{1pt}
    \rowcolor{gray!10}
    \textbf{Dataset} & \textbf{Simulation Objectives} & \textbf{Input Attributes} & \textbf{Target Attributes} \\
    \hline
    GSS & Public Opinions & \textit{background}, sample size, batch size, \uline{distribution (military, crime, education)} & natenvir, natheal, natcity \\
    \hline
    Trell SMU & Media Usage Habit & \textit{background}, sample size, batch size, \uline{distribution (age group, city tier, duration, creation, averageCompletion)} & \makecell[l]{content\_views \\ weekends\_trails\_watched\_per\_day \\ weekdays\_trails\_watched\_per\_day} \\
    \hline
    YPS & Music \& Movie Preferences & \textit{background}, sample size, batch size, \uline{distribution (Music, Slow songs or fast songs, Dance, Folk, Country, Classical music, Musical, Pop, Rock, Metal or Hardrock, Punk, Hiphop, Rap, Reggae, Ska, Swing, Jazz, Rock n roll, Alternative, Latino, Techno, Trance, Opera, Movies, Horror, Romantic, Sci-fi, War, Fantasy/Fairy tales, Animated)} & Action, Documentary, Thriller, Comedy \\
    \hline
    RECS & Household Fuel Usage & \textit{background}, sample size, batch size, \uline{distribution (Gender (HHSEX), Age Range, Employment Status, State (State-Postal))} & KWH, TOTALDOL \\
    \Xhline{1pt}
  \end{tabular}
  }
\end{table}

\begin{table}[htbp]
  \scriptsize
  \setlength{\tabcolsep}{2pt}
  \renewcommand{\arraystretch}{1.5}
  \caption{Detailed Performance of LLMs on prediction tasks in the FAS scenario.}
  \label{tab:fas_perf_detail}
  \resizebox{\textwidth}{!}{
  \begin{tabular}{l|l|l|l|l|l|l|l|l|l}
    \Xhline{1pt}
    \multirow{2}{*}{\textbf{Dataset}} & \multirow{2}{*}{\textbf{Target Attribute}} &
    \multicolumn{4}{c|}{\textbf{Zero-Context Generation}} &
    \multicolumn{4}{c}{\textbf{Context-Enhanced Generation}} \\
    \cline{3-6} \cline{7-10}
    & & GPT-3.5 Turbo & GPT-4 Turbo & LLaMA 3.0 & LLaMA 3.1 & GPT-3.5 Turbo & GPT-4 Turbo & LLaMA 3.0 & LLaMA 3.1 \\
    \hline
    \multirow{3}{*}{GSS}
      & natenvir & \textbf{0.6555} & 0.6495 & 0.6449 & 0.6487 & 0.6504 & \textbf{0.7005} & 0.6649 & 0.6812 \\
      \cline{2-10}
      & natheal & 0.6526 & 0.6517 & 0.6477 & \textbf{0.6533} & 0.6683 & \textbf{0.6955} & 0.6635 & 0.6777 \\
      \cline{2-10}
      & natcity & 0.6388 & 0.6457 & \textbf{0.6596} & 0.6540 & 0.6520 & \textbf{0.6624} & 0.6430 & 0.6589 \\
    \hline
    \multirow{3}{*}{Trell SMU}
      & content\_views & 0.2992 & \textbf{0.3455} & 0.2772 & 0.3330 & 0.4160 & \textbf{0.4276} & 0.3219 & 0.3066 \\
      \cline{2-10}
      & weekends\_trails\_... & 0.2848 & 0.2822 & 0.1823 & \textbf{0.2921} & \textbf{0.2797} & 0.2732 & 0.2025 & 0.2497 \\
      \cline{2-10}
      & weekdays\_trails\_... & 0.3669 & \textbf{0.3683} & 0.2463 & 0.2724 & 0.3272 & \textbf{0.3844} & 0.2426 & 0.2814 \\
    \hline
    \multirow{4}{*}{YPS}
      & Action & 0.6525 & \textbf{0.6968} & 0.6478 & 0.6368 & 0.6526 & \textbf{0.6968} & 0.6478 & 0.6368 \\
      \cline{2-10}
      & Documentary & 0.6472 & \textbf{0.6968} & 0.6599 & 0.6542 & 0.6472 & \textbf{0.6968} & 0.6599 & 0.6542 \\
      \cline{2-10}
      & Thriller & 0.6650 & \textbf{0.7045} & 0.6385 & 0.6407 & 0.6650 & \textbf{0.7045} & 0.6385 & 0.6407 \\
      \cline{2-10}
      & Comedy & 0.7409 & \textbf{0.7859} & 0.6815 & 0.6837 & 0.7409 & \textbf{0.7859} & 0.6815 & 0.6837 \\
    \hline
    \multirow{2}{*}{RECS}
      & KWH & 0.6208 & \textbf{0.6267} & 0.5484 & 0.3753 & 0.5976 & \textbf{0.6302} & 0.5763 & 0.5113 \\
      \cline{2-10}
      & TOTALDOL & 0.6460 & \textbf{0.6501} & 0.5709 & 0.3900 & 0.6209 & \textbf{0.6504} & 0.6003 & 0.5946 \\
    \Xhline{1pt}
  \end{tabular}
  }
\end{table}

\begin{figure}[htbp]
\centering
\includegraphics[width=0.7\textwidth]{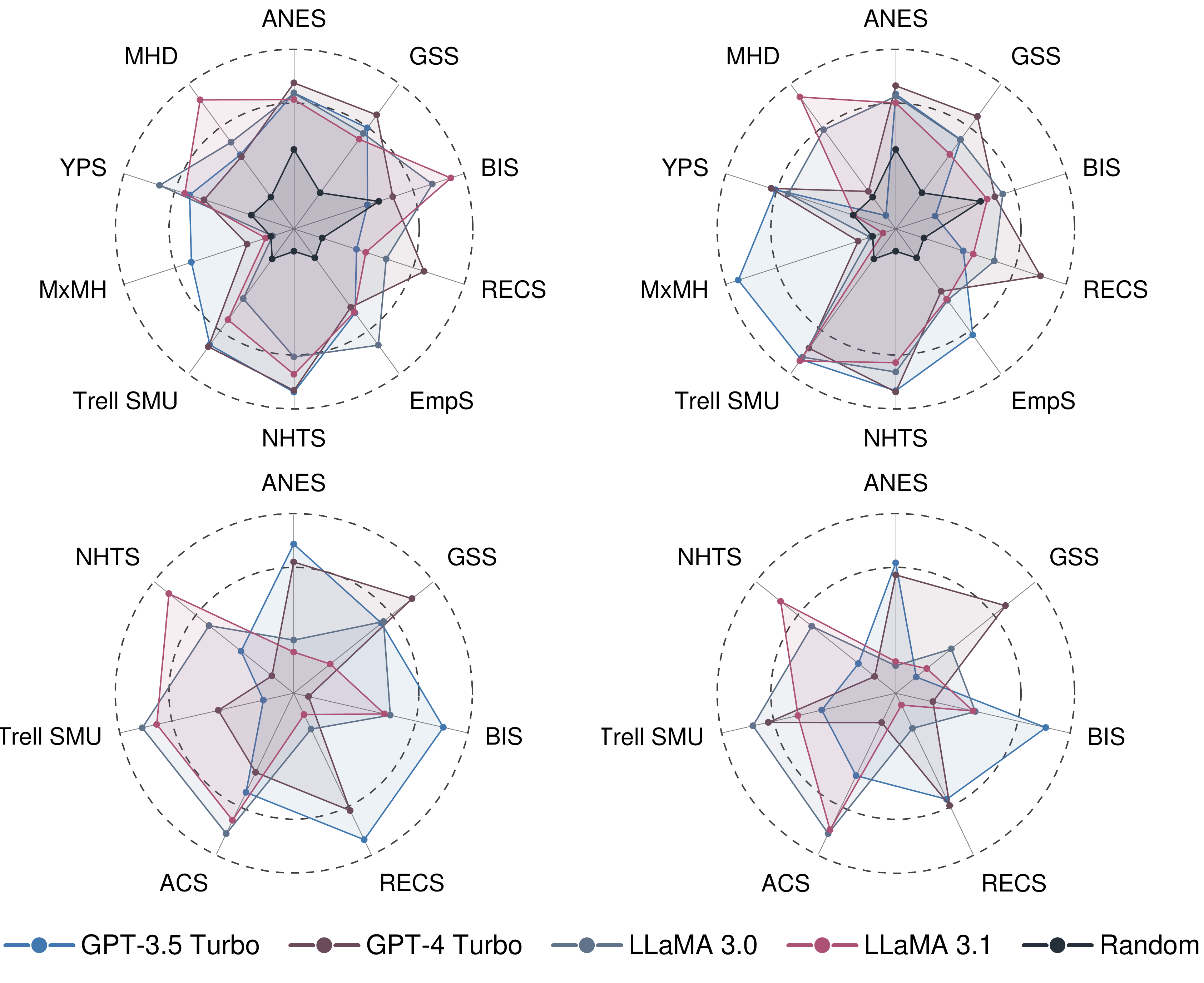}
\caption{Radar plots illustrating model performance on the PAS scenario across datasets (each axis is individually scaled per dataset to enhance the visibility of relative differences). 
The top row shows multiple-choice tasks, and the bottom row shows numerical prediction, under zero-shot evaluation (left) and few-shot evaluation (right). 
Higher values indicate better performance.}
\label{fig:pas_radar}
\end{figure}

\begin{figure}[htbp]
\centering
\includegraphics[width=0.7\textwidth]{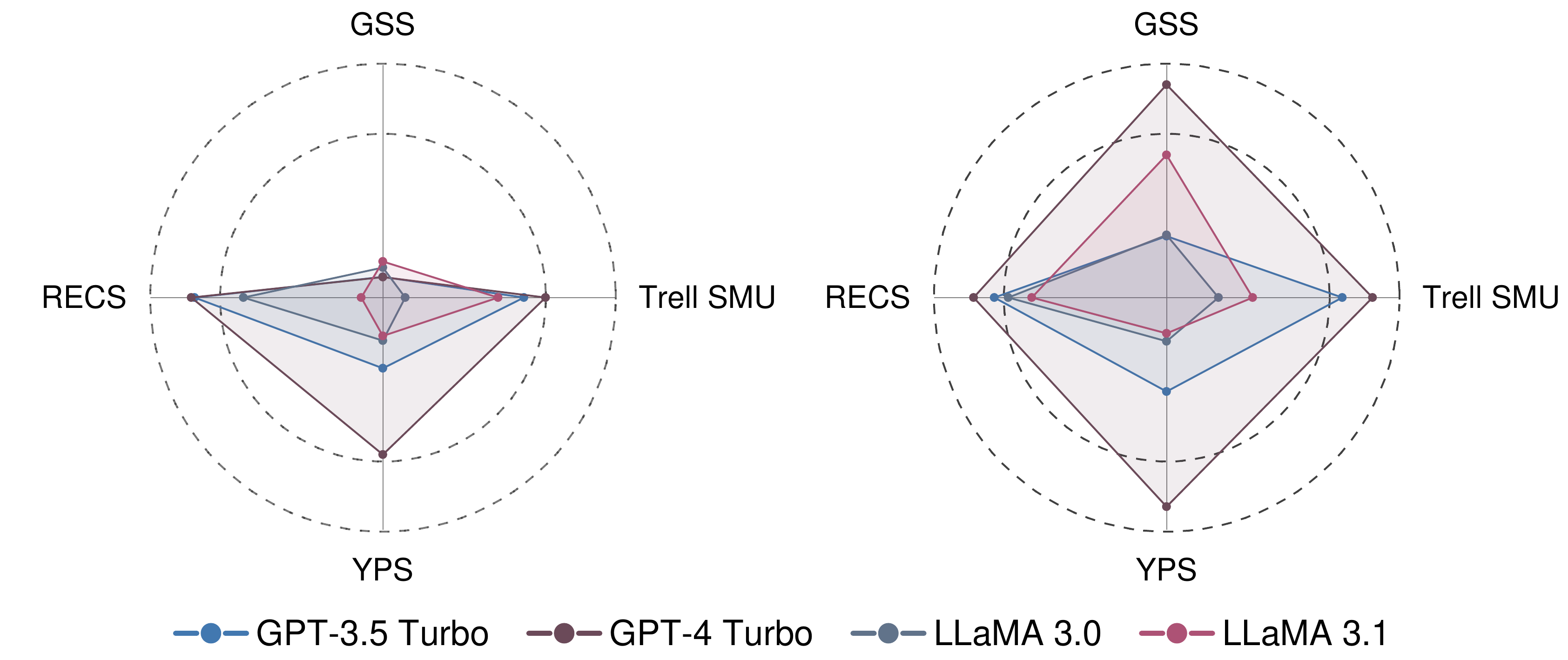}
\caption{Radar plots illustrating model performance on the FAS scenario across datasets (each axis is individually scaled per dataset to enhance the visibility of relative differences). 
Results are shown under zero-context generation (left) and context-enhanced generation (right). 
Higher values indicate better performance.}
\label{fig:fas_radar}
\end{figure}

To provide a clearer and more intuitive overview for comparing model performance across datasets, prediction tasks, and evaluation strategies in PAS and FAS, Figure~\ref{fig:pas_radar} (PAS) and Figure~\ref{fig:fas_radar} (FAS) display radar plots with axis values normalized within each dataset, facilitating clearer visualization of relative performance differences across models and tasks. 
Across both scenarios, prediction tasks are divided into two types: multi-choice (with several options) and numerical (involving continuous value estimation). 
Evaluation metrics include accuracy for multi-choice tasks and a normalized reciprocal (i.e., monotonic transformation) of KL divergence for numerical tasks. 
Together, these metrics provide a quantitative basis for comparing model performance across diverse tasks. 
For detailed definitions and computational procedures, please refer to Section~\ref{Metrics}. 

To ensure the robustness and reproducibility of our findings, each evaluation trial in the PAS scenario was conducted with three independent runs under the same conditions. 
The reported results present both the average values and standard deviations across repetitions, which helps assess the stability and reliability of model performance. 
In contrast, FAS adopts a design where each run consists of multiple prompt batches. Each batch generates several data outputs, and the aggregation of these outputs across all batches forms the basis for the final performance evaluation. 
This structure yields many data samples within a single run, offering sufficient variation for a stable and representative assessment. Hence, additional independent runs were unnecessary for FAS, as internal variability already offers a reliable performance basis.

\subsection{Prompt Instances}

\subsubsection{PAS}
In the PAS scenario, two evaluation strategies are employed: zero-shot evaluation and few-shot evaluation. 
The key distinction lies in that the few-shot setting offers several example attribute–answer pairs as demonstrations, intending to enhance the LLM’s understanding and improve its predictive performance. 
As shown in Figure~\ref{fig:prompt_design_pas}, the example prompt distinction highlights the shift in the amount of information provided to the model.

\subsubsection{FAS}
In the FAS scenario, we aim to generate group-level data that reflects the distribution of attributes within a specified population. 
This scenario employs two generation paradigms: zero-context and context-enhanced generation. In the zero-context generation, the model generates data without any additional context about the specific population. In contrast, the context-enhanced generation incorporates distributional information, such as the proportions of gender, age, and other attributes within the dataset, which helps the model generate more realistic and representative data. Figure~\ref{fig:prompt_design_fas} illustrates an example of these two generation paradigms, showing how the prompt structure changes depending on whether context is provided or not.

\FloatBarrier 
\onecolumn
\begin{figure}[H]
    \centering  
    \scalebox{1}{
        \begin{tcolorbox}[
            colframe=black,          
            colback=gray!5,          
            coltitle=white,          
            boxrule=0.5mm,          
            colbacktitle=black,     
            width=1\textwidth,       
            arc=5mm,                 
            title={\centering\small \textbf{Partial Attribute Simulation}}  
        ]
            \scriptsize
            {\small\bfseries \textcolor{black}{Background Description}} \\
            The survey content was shaped to a significant degree by ideas offered by the ANES user community through public solicitations and by members of the ANES National Advisory Board. Additional questions were included to capture changes in the political environment and recent developments in social science research. The questionnaire includes questions about voting behavior, candidate traits, political engagement, ideological orientations, racial identity and stereotyping, and many topical issues, including \#MeToo, immigration, impeachment, and the coronavirus pandemic.
            \vspace{1em}
            
            {\small\bfseries \textcolor{black}{Respondent Profiles}} \\
            This person is 40 years old, a homeowner, and has an income between \$5,000 and \$9,999. He has a voting tendency to select someone to vote for, but their summarized participation score is 0.
            
            \vspace{1em}

            {\small\bfseries \textcolor{black}{Conditions}} \\
            This person's participation in political activities is as follows:  
            \begin{itemize}
                \item \textit{Has not attended a meeting to discuss political or social concerns}.
                \item \textit{Has not donated to an organization focused on a political or social issue.}
                \item \textit{Has not joined a protest march, rally, or demonstration.}
                \item \textit{Has not posted messages or comments online about political issues or campaigns.}
                \item \textit{Has not tried to persuade anyone to vote in any direction.}
                \item \textit{Has not worn a campaign button, placed a sticker on their car, or displayed a sign in their window or yard.}
                \item \textit{Has a cold or unfavorable feeling (score 15.0, in the range of [0,20]) towards Mike Pence and Andrew Yang.}
                \item \textit{Has a cold or unfavorable feeling (score 0.0, in the range of [0,20]) towards Nancy Pelosi, Marco Rubio, Alexandria Ocasio-Cortez, and Nikki Haley.}
            \end{itemize}
            
            \vspace{1em}
            
            {\small\bfseries \textcolor{black}{Task Demands}} \\
            The data come from the American National Election Studies 2020 (ANES2020). As a sociologist and political scientist, you need to analyze the public opinion of citizens in this dataset and predict their possible election attitudes. Based on the information provided, can you help decide how this person would rate Clarence Thomas?

            \begin{tcolorbox}[colback=cyan!10, colframe=white, boxrule=0mm, arc=5mm, width=\textwidth-0cm]
            {\small\bfseries \textcolor{blue!50!black}{Few-Shot Examples (Zero-Shot Not Used)}} \\
            Here are some examples and corresponding answers:\\
            \textit{\textbf{Example 1:}}
                \begin{itemize}\setlength{\itemsep}{0pt}
                    \item \textit{This person is 69 years old.}
                    \item \textit{This person is a homeowner.}
                    \item \textit{This person has an income between \$10,000 and \$14,999.}
                    \item \textit{This person probably would not vote.}
                    \item \textit{This person has a summarized participation score of 0.}
                    \item \textit{This person has a neutral feeling (score 60.0, in the range of [40, 60]) towards Mike Pence, Andrew Yang, Dr. Anthony Fauci, Joe Biden, and Donald Trump.}
                    \item \textit{\textbf{The answer is 60.}}
                \end{itemize}
            \textit{\textbf{Example 2:}} ... \\
            \textit{\textbf{Example 3:}} ...
            \end{tcolorbox}

            \vspace{1em}
            {\small\bfseries \textcolor{black}{Answer Format}} \\
            Your response should consist of just one number between [0, 100] to reflect the person's attitude, without any additional text, explanation, or even a space character. Here is an example of the required response format that you should follow:  \begin{itemize}\setlength{\itemsep}{0pt}
                \item \textit{If you think this person has a slightly warm or favorable feeling, you should respond with JUST an integer number, such as 62.} 
                \item \textit{For more or less favorable feelings, the number may be larger or smaller. }
                \item \textit{\textbf{So your response should be like: 62.}}
            \end{itemize}
        \end{tcolorbox}
    }
    \captionof{figure}{A representative prompt utilized in zero-shot and few-shot evaluation conducted on the ANES dataset within the PAS scenario.}
    \label{fig:prompt_design_pas}  
\end{figure}
\FloatBarrier 
\clearpage
\onecolumn
\begin{figure}[H]
    \centering  
    \scalebox{1}{
        \begin{tcolorbox}[
            colframe=black,          
            colback=gray!5,          
            coltitle=white,          
            boxrule=0.5mm,          
            colbacktitle=black,     
            width=1\textwidth,       
            arc=5mm,                 
            title={\centering\small \textbf{Full Attribute Simulation}}  
        ]
            \scriptsize
            {\small\bfseries \textcolor{black}{Background Description}} \\
            The Residential Energy Consumption Survey (RECS), conducted by the U.S. Energy Information Administration (EIA), is a nationally representative study designed to collect detailed information on household energy usage, expenditures, and related demographics. Since its inception in 1978, RECS has played a vital role in providing insights into energy consumption patterns across various housing units in the United States. The 2015 RECS dataset, for instance, surveyed over 5,600 households, representing approximately 118.2 million primary residences. The survey encompasses data on energy sources such as electricity, natural gas, propane, and fuel oil, along with expenditure estimates for heating, cooling, and other end uses. This dataset has been instrumental in energy policy analysis, efficiency improvements, and forecasting future consumption trends, making it a cornerstone of energy research in the residential sector.
            
            \vspace{0.5em}

            
            \begin{tcolorbox}[colback=cyan!10, colframe=white, boxrule=0mm, arc=5mm, width=\textwidth-0cm]
            {\small\bfseries \textcolor{blue!50!black}{Zero-Context Generation}} \\
            You are a data scientist and socioeconomic analyst. Your task is to generate a synthetic dataset simulating residential energy consumption and household characteristics based on general knowledge and logical assumptions. No prior distributional information or feature statistics will be provided. Use your expertise to model realistic energy usage patterns, household demographics, and expenditure data for a nationally representative set of housing units.
            \vspace{0.5em}
            
            {\small\bfseries \textcolor{black}{Prediction Question}} \\
            Based on your understanding, prior knowledge, and general reasoning, predict the following for a household: 
            \begin{itemize}
                \item What is the total electricity use \textit{(KWH)} of this household in 2020?  
                \item What is the total electricity cost \textit{(DOLLAREL)} of this household in 2020?  
                \item What is the total energy cost \textit{(TOTALDOL)} of this household in 2020? 
            \end{itemize}
            \vspace{0.5em}

            {\small\bfseries \textcolor{black}{Sample Size, Batch Size, and Answer Format}} \\
            Generate 20 rows of data in each batch, and repeat this process for 100 batches to simulate a comprehensive dataset. For each batch, provide the predictions directly in the following format: [\textit{KWH}, \textit{DOLLAREL}, \textit{TOTALDOL}]. 
            After completing a batch, explicitly indicate the batch number before moving to the next batch (e.g., "Batch 1 complete"). Continue this process until all 100 batches are generated. The format must be a JSON string representing a three-dimensional array. Also, make sure that it is an array of arrays with no objects, like in a spreadsheet. After generating, only show the data you generated without additional words. Remember, the records should closely reflect the RECS dataset.
            \end{tcolorbox}
            
            \vspace{1em}
        
            \begin{tcolorbox}[colback=pink!30, colframe=white, boxrule=0mm, arc=5mm, width=\textwidth-0cm]
            {\small\bfseries \textcolor{blue!50!black}{Context-Enhanced Generation}} \\
            You are a data scientist and socioeconomic analyst. I will provide you with the demographic distribution of a population, including the proportions of gender \textit{(HHSEX)}, age groups \textit{(HHAGE)}, employment status \textit{(EMPLOYHH)}, and state location \textit{(state\_postal)}. Based on these distributions, your task is to generate a synthetic dataset that reflects the specified demographic structure. The generated dataset should include plausible and consistent combinations of features that align with the given proportions, simulating a realistic social survey dataset. Ensure that the dataset captures meaningful patterns and relationships between features commonly observed in real-world surveys. 
            \vspace{0.5em}
            
            {\small\bfseries \textcolor{black}{Distribution Information}} \\
            You may be provided with the following distribution data: 
            \begin{itemize}[leftmargin=10mm]
                \item {\textit{HHSEX (sex)}}: {[Female: 0.5408, Male: 0.4592]}
                \item {\textit{HHAGE (age group)}}: {[18-30: 0.0868, 31-40: 0.1503, 41-50: 0.1496, 51-60: 0.1848, 61-70: 0.2087, 71-80: 0.1574, 81-90: 0.0623]}
                \item {\textit{EMPLOYHH (employment status)}}: {[Employed full-time: 0.4706, Employed part-time: 0.0868, Not employed: 0.1166, Retired: 0.3261]}
                \item {\textit{state\_postal (state location)}}: {[Alabama: 0.0131, Alaska: 0.0168, Arkansas: 0.0145, California: 0.0623, Colorado: 0.0195, Connecticut: 0.0159, Delaware: 0.0077, ..., Wisconsin: 0.0193, Wyoming: 0.0103]
                }
            \end{itemize}
            \vspace{0.5em}
        
            {\small\bfseries \textcolor{black}{Prediction Question}} \\
            Based on the provided distributions of gender \textit{(HHSEX)}, age range \textit{(HHAGE)}, employment status \textit{(EMPLOYHH)}, and state location \textit{(state\_postal)}, predict the total electricity use \textit{(KWH)}, the total electricity cost \textit{(DOLLAREL)}, and the total energy cost \textit{(TOTALDOL)} for each household in 2020. \\

            {\small\bfseries \textcolor{black}{Sample Size, Batch Size, and Answer Format}} \\
            (Same as Zero-Context Generation)
            
            \end{tcolorbox}
        \end{tcolorbox}
    }
    \captionof{figure}{A representative prompt utilized in zero-context and context-enhanced generation conducted on the RECS dataset within the FAS scenario.}  
    \label{fig:prompt_design_fas}  
\end{figure}

\FloatBarrier 
\clearpage

\end{document}